\documentclass[journal,twoside,web]{ieeecolor}
\usepackage{tmi}
\usepackage{amsmath}
\usepackage{mathtools}
\usepackage{multirow}
\usepackage{array}

\usepackage{calc}
\usepackage{tikz}
\usetikzlibrary{calc}
\usepackage{overpic}

\usepackage[colorlinks,
							urlcolor=blue,
							citecolor=blue,
							implicit=false
						]{hyperref}
\usepackage{etoolbox}

\makeatletter
\@ifundefined{color@begingroup}%
{\let\color@begingroup\relax
	\let\color@endgroup\relax}{}%
\def\fix@ieeecolor@hbox#1{%
	\hbox{\color@begingroup#1\color@endgroup}}
\patchcmd\@makecaption{\hbox}{\fix@ieeecolor@hbox}{}{\FAILED}
\patchcmd\@makecaption{\hbox}{\fix@ieeecolor@hbox}{}{\FAILED}
\renewcommand{\maketag@@@}[1]{\hbox{\m@th\normalsize\normalfont#1}}%
\makeatother

\def\BibTeX{{\rm B\kern-.05em{\sc i\kern-.025em b}\kern-.08em
		T\kern-.1667em\lower.7ex\hbox{E}\kern-.125emX}}
\graphicspath{{fig/},{../../pami/delineation/fig/}}

\newcommand{\baseline}[0]{{\it OrigAnnot}}
\newcommand{\snakeFull}[0]{{\it SnakeFull}}
\newcommand{\snakeFast}[0]{{\it SnakeFast}}
\newcommand{\snakeSimple}[0]{{\it SnakeSimple}}
\newcommand{\CCQ}[0]{{\it CCQ}}
\newcommand{\APLS}[0]{{\it APLS}}
\newcommand{\TLTS}[0]{{\it TLTS}}

\newcommand{\neurons}[0]{{\it Brain}}
\newcommand{\neuronsmisal}[0]{{\it Neurons}}
\newcommand{\mra}[0]{{\it MRA}}
\newcommand{\synth}[0]{{\it Synthetic}}
\newcommand{\UNet}[0]{{\it UNet}}
\newcommand{\DRU}[0]{{\it DRU}}
\newcommand{\DS}[0]{{\it DS6}}
\newcommand{\COPLE}[0]{{\it NR-Dice}}
\newcommand{\QAM}[0]{{\it QAM}}
\newcommand{\MSE}[0]{{\it L2}}
\newcommand{\MAE}[0]{{\it L1}}

\newcommand{\node}{v}

\DeclareMathOperator*{\argmin}{arg\,min}

\newcommand{\mycomment}[1]{}

\newif\ifdraft
\draftfalse
\usepackage{booktabs, multirow}

\definecolor{orange}{rgb}{1,0.5,0}
\definecolor{violet}{RGB}{70,0,170}
\definecolor{magenta}{RGB}{170,0,170}
\definecolor{dgreen}{RGB}{0,150,0}

\usepackage[normalem]{ulem}

\ifdraft
 \newcommand{\PF}[1]{{\color{red}{\bf PF: #1}}}

 \newcommand{\MS}[1]{{\color{dgreen}{\bf MS: #1}}}
 
 \newcommand{\MK}[1]{{\color{magenta}{\bf MK: #1}}}
 \newcommand{\mk}[1]{{\color{magenta} #1}}
 \newcommand{\change}[1]{{\color{dgreen} #1}}
\else
 \newcommand{\PF}[1]{}
 
 \newcommand{\PUW}[1]{}
 
 \newcommand{\MS}[1]{}
 
 \newcommand{\MK}[1]{}
 \newcommand{\mk}[1]{#1}
 \newcommand{\change}[1]{{#1}}
\fi

\newcommand{\comment}[1]{}

\newcommand{\bA}{\mathbf{A}}
\newcommand{\bC}{\mathbf{C}}
\newcommand{\bI}{\mathbf{I}}

\newcommand{\bX}{\mathbf{X}}

\newcommand{\bc}{\mathbf{c}}

\newcommand{\by}{\mathbf{y}}

\newcommand{\mC}{\mathcal{C}}

\begin{document}
\title{Adjusting the Ground Truth Annotations for Connectivity-Based Learning to Delineate}

\author{Doruk Oner, Mateusz Kozi\'{n}ski, Lenoardo Citraro, and Pascal Fua, \IEEEmembership{Fellow, IEEE}
\thanks{
This work was supported in part by the Swiss National Science Foundation Sinergia grant no.\ 177237
and by the FWF Austrian Science Fund Lise Meitner grant no.\ M3374.}
\thanks{
Doruk Oner, Leonardo Citraro, and Pascal Fua are with the Computer Vision Laboratory, EPFL, Switzerland (mail:\{doruk.oner, leonardo.citraro, pascal.fua\}@epfl.ch). Mateusz Kozinski is with the Institute of Computer Vision and Graphics, TU Graz, Austria (mail:mateusz.kozinski@icg.tugraz.at).}
}
\maketitle % typeset the header of the contribution
%

%
%
% !TEX root = ../top.tex
% !TEX spellcheck = en-US

\begin{abstract}

Deep learning-based approaches to delineating 3D structure depend on accurate annotations to train the networks. Yet in practice, people, no matter how conscientious, have trouble precisely delineating in 3D and on a large scale, in part because the data is often hard to interpret visually and in part because the 3D interfaces are awkward to use. 

In this paper, we introduce a method that explicitly accounts for annotation inaccuracies. 
To this end, we treat the annotations as active contour models that can deform themselves while preserving their topology. This enables us to jointly train the network and correct potential errors in the original annotations. The result is an approach that boosts performance of deep networks trained with potentially inaccurate annotations. Code has been released at \href{https://github.com/doruk-oner/AdjustingAnnotationswithSnakes}{https://github.com/doruk-oner/AdjustingAnnotationswithSnakes}.

\end{abstract}


% !TEX root = ../top.tex
% !TEX spellcheck = en-US

\section{Introduction}

As in many areas of computer vision, deep networks now deliver state-of-the-art results for delineation tasks, such as finding axons and dendrites in 3D light microscopy images. However, their performance depends critically on the accuracy of the ground-truth data used to train them. This is especially true when the delineation task is treated as a segmentation one and the network is trained by minimizing the cross-entropy between the centerline predictions and ground-truth annotations, which is one of the most popular paradigms.

In practice, these so-called ground-truth annotations are usually supplied manually by an annotator who may not draw with the utmost accuracy and can therefore easily be a few voxels off the true centerline. This is not a matter of carelessness but a consequence of 3D delineation being truly difficult to do well on a large scale. As a result, inaccurate annotations are more the rule than the exception and this adversely affects how well the networks ultimately perform. One solution would be to have several annotators delineate the same data and combine their delineations. However, this would turn an already tedious, slow, and expensive process into an even slower and more expensive one that almost no one can afford. 

% !TEX root = ../top.tex
% !TEX spellcheck = en-US

\begin{figure}
\centering
\includegraphics[width=\columnwidth]{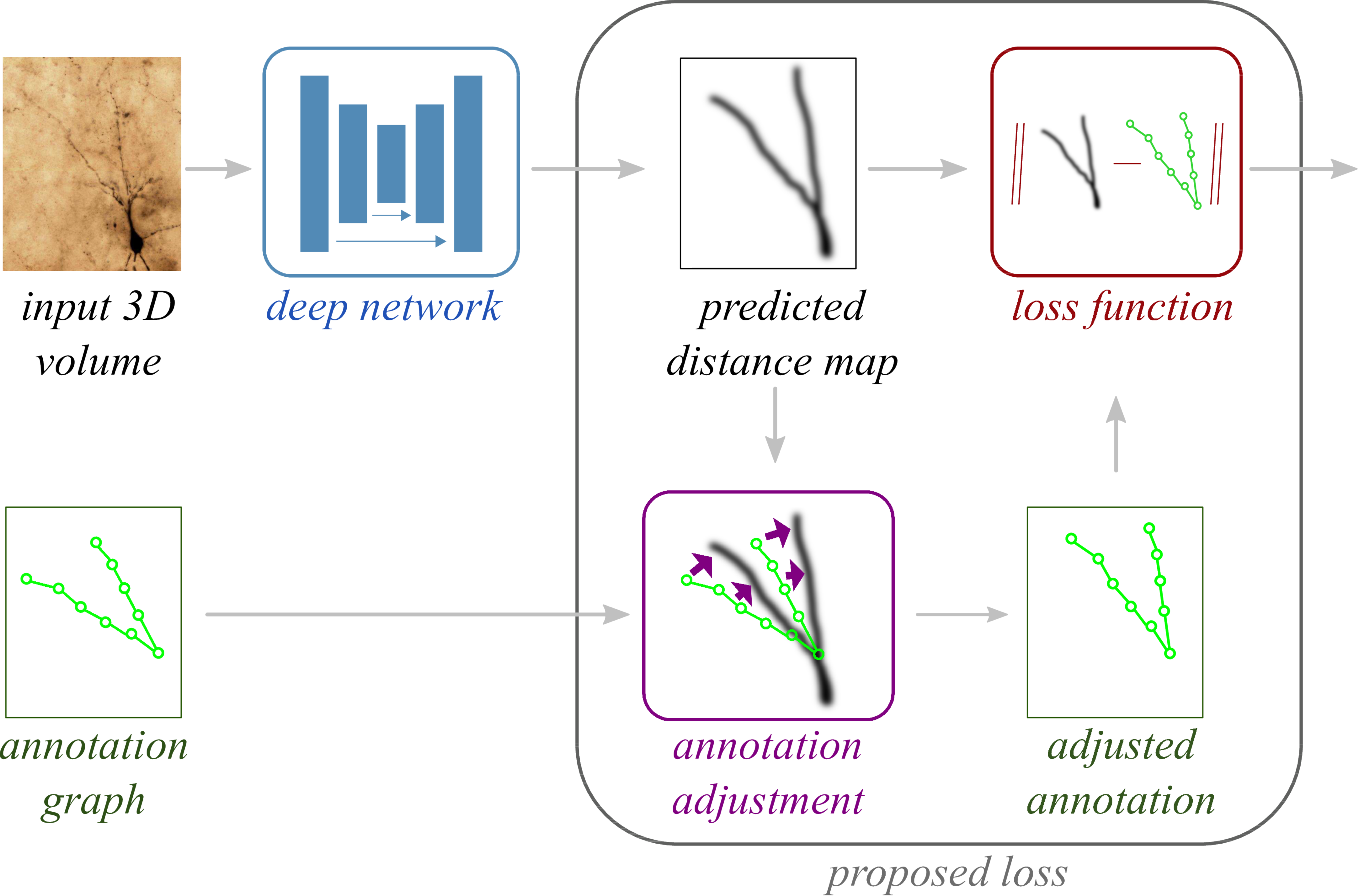}
\caption{
{\bf Our approach.} 
To account for annotation inaccuracies during training,  we jointly train the network and adjust the annotations while preserving their topology.
}
\label{fig:teaser}
\end{figure}

% !TEX root = ../top.tex
% !TEX spellcheck = en-US

\begin{figure}[!h]
{
\small
\setlength{\tabcolsep}{1pt}
\begin{tabular}{cccc} 
\includegraphics[width=0.24\columnwidth]{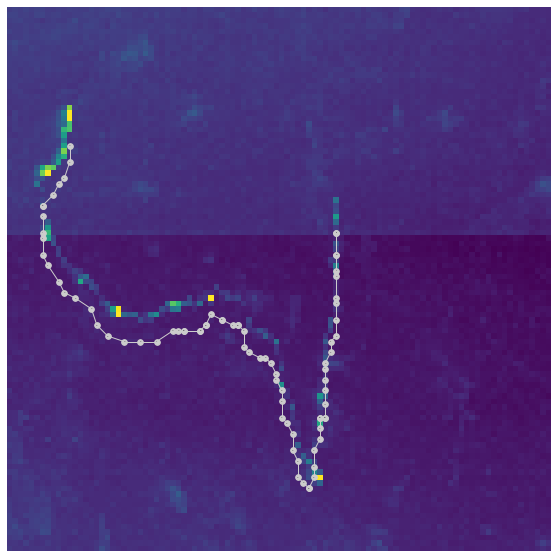} &
\begin{overpic}[width=0.24\columnwidth,]{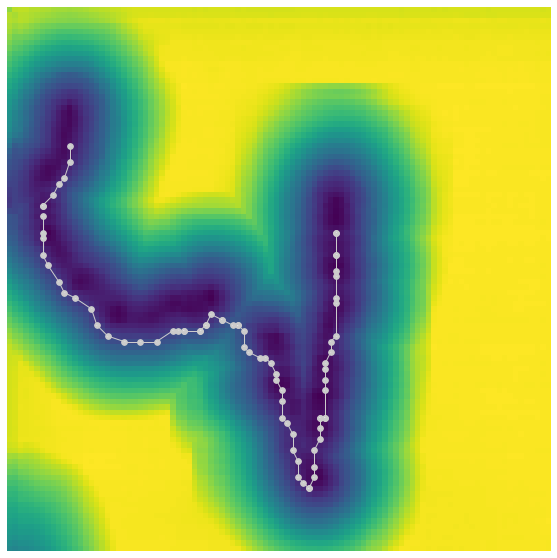}\put(19,18){\color{red}\vector(1,1){20}}\end{overpic} &
\includegraphics[width=0.24\columnwidth]{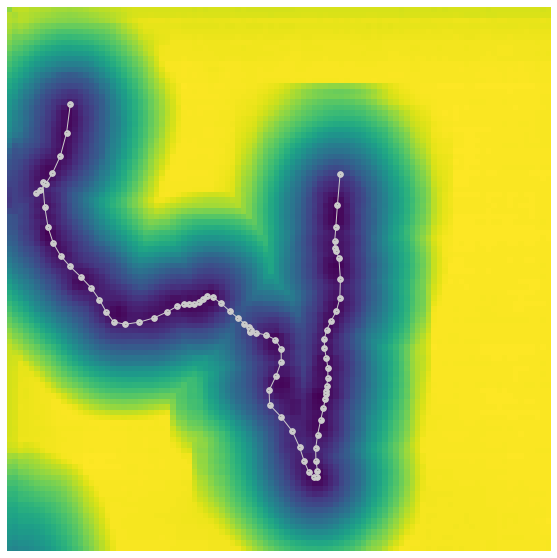} &
\includegraphics[width=0.24\columnwidth]{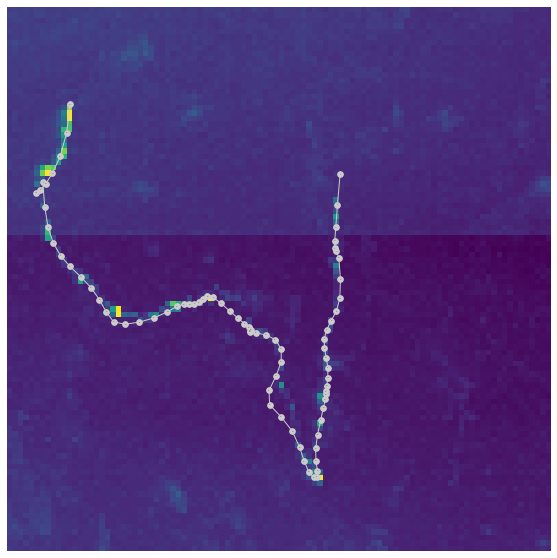} \\
(a) & (b) & (c) & (d)
\end{tabular}
}
\vspace{-2mm}
\caption{
\label{fig:teaser2}
{\bf Correcting an inaccurate annotation.}
(a) A microscopy scan of a neurite with an inaccurate annotation overlaid in white. (b) Distance map predicted by the deep net. Ideally, the pixels crossed by the centerline should have value zero (dark color).
In practice, this is not always the case.
There are non-zero values in the area indicated by the red arrow, presumably because the neurite is hardly visible there. Nevertheless, the distance map is sufficiently good to adjust the annotation. The adjusted annotation is shown in (c) and (d). This network retrained with adjusted annotations can now generate a better distance map even where the neurite is barely visible.
}
\end{figure}

In this paper, we introduce a method that explicitly accounts for annotation inaccuracies and delivers the same performance as if they were perfectly accurate. Our main insight is that the annotations are usually imprecise more in terms of the 3D location of the centerlines than of the topology of the graph they define. We can therefore treat them as deformable contours forming a graph that can be refined by moving its nodes while preserving its structure. We cast this approach to training a deep network as a joint optimization over the network parameters and node positions. We then show that we can eliminate the node variables from the optimization problem, which can then be solved by minimizing a loss function. This loss function accounts for the annotation's lack of spatial precision. It can be minimized in the traditional manner and the output of the re-trained network used to refine the annotation. 

Fig.~\ref{fig:teaser} depicts our approach and Fig.~\ref{fig:teaser2} showcases its behavior. We will demonstrate that it brings substantial improvements when training networks to delineate neurons in two-photon and confocal microscopy image stacks. Hence, our contribution is an automated approach to better leveraging inaccurate training data, which, in our experience, represents the vast majority of data available to practitioners.

% !TEX root = ../top.tex
% !TEX spellcheck = en-US

\section{Related work}
\label{sec:related}

\subsection{Automated Delineation}

Automatic delineation of curvilinear structures has been an active research topic for decades. It has evolved from manually designing filters that respond strongly to tubular structures~\cite{Frangi98,Law08,Turetken13c} to feeding hand-designed features into boosted trees~\cite{Wu12a,Breitenreicher13}, support vector machines~\cite{Huang09}, or GradientBoost~\cite{Sironi16a}, and finally to fully relying on neural networks~\cite{Mnih10,Ganin14a,Maninis16,Peng17,Mosinska20}.

The latter now routinely deliver the best  performance when properly trained. However, obtaining accurately annotated data, especially in 3D, is a challenge. In practice it is rarely available in sufficient quantities. And what annotated data there is, is rarely accurate because manually delineating 3D structures is challenging. Introducing a degree of self-supervision is a way to address this difficulty~\cite{Bengio13a,Doersch17} but this does not detract from the fact that the training would work even better if the available annotated data were accurate.  This can be partially ascribed to the fact that most current networks are trained by minimizing the standard cross entropy or differentiable intersection-over-union loss~\cite{Mattyus17}. As pixel-wise measures, both are sensitive to even small misplacements of the linear structures' centerlines. In~\cite{Mosinska18}, this is partially addressed by introducing a loss component that accounts for global statistics of the network output, but the cross entropy remains a key component of the overall loss. Similarly, the method of~\cite{Oner21a} relies on introducing a topology-preserving term but still depends on the annotation being accurate.

Accuracy can be improved by having several people annotate and combining their results using robust statistics. This is effective but even more expensive than obtaining one set of annotations and therefore out of reach for most practitioners.  The problem can be partially alleviated by annotating only in 2D projections of the 3D data volumes~\cite{Peng14,Zhou16,Kozinski20}, which is easier, but may result in even less precise annotations than those performed in 3D. 

A similar problem to the one we address here also arises in the context of two-dimensional semantic boundary detection. The outlines one finds in annotated training sets are often rather imprecise and training the networks to nevertheless discover contours that overlap with them is an issue. In~\cite{Yu18d}, training is reformulated as simultaneously optimizing the parameters of a deep net and correcting the annotations by solving a mixed binary-continuous optimization problem. However, unlike in our approach, preservation of annotation topology is not warranted and the corrections may break the continuity of annotations. This is a major problem when tracing neurons or blood vessels, because topology changes influence the biological interpretation of the results. The same problem is addressed in~\cite{Acuna19} by proposing a neural layer and a loss function that can be added on top of an edge detector and make it possible to find more accurate contours than those in the annotations. However, because the regions are represented in an implicit fashion, there is no more guarantee than in~\cite{Yu18d} that the annotations' connectivity will be preserved. Connectivity being at the heart of our applications, we therefore chose to use explicit deformable models, such as those described below. 

\subsection{Handling Noisy Annotations}

{
Even though we know of no other algorithm that adjusts the geometry of centerline annotations during training, explicitly accounting for the fact that the annotations are noisy has received some attention. In~\cite{Wang20i}, annotations produced by non-expert annotators are accommodated by means of a dedicated distillation architecture and a noise-robust Dice loss. In~\cite{Chatterjee20},  a dedicated network architecture and a semi-supervised training routine encourage equivariance to deformations to handle potential inaccuracies resulting from using a heuristic annotation tool. In~\cite{Zhu19}, annotation noise is handled by a quality assessment module that discounts the loss in regions where the estimated label quality is low. Similarly, in~\cite{Min19}, a distillation training setup and architecture based on self-attention are used to suppress the influence of erroneous labels on the trained network. In contrast to all these approaches, ours explicitly distinguishes between inaccuracies in position and topological errors. Because the former occur far more frequently than the latter, our loss function adjusts the centerline locations, while preserving the topology of the annotations. 
}

\subsection{Deformable Contour Models}

Deformable contours~\cite{Kass88,Terzopoulos88,Fua90} were initially introduced as a means to semi-automatically delineate simple contours while imposing smoothness constraints on the resulting outlines. They were later generalized to model network structures~\cite{Fua96f,Butenuth12} that can deform while preserving their topology. They are therefore well suited for refining our inaccurate annotations under the assumption they are topologically correct but that their locations are imprecise. 

More recent deformable contours rely on minimizing energy functions generated by deep networks~\cite{Marcos18,Cheng19,Wang20i,Hatamizadeh20}, which enables end-to-end learning. Unlike in these methods, which rely on evolving the contour for segmenting the image at test time, our use of deformable contours is limited to adjusting the annotations during training.

Active appearance models~\cite{Cootes01} enable modelling the appearance of imaged objects, in addition to their shape. They can be learnt from coarse annotations, which are adjusted when fitting the model to the data~\cite{Ramnath08}. The level of detail of the active appearance model can then be increased and, before the more detailed model is fitted to the data, it can be initialized with the parameters of its less detailed version. In this work, we also adjust the annotation during learning, but represent them as network snakes, and train a deep convolutional network, instead of fitting an active appearance model.

% !TEX root = ../top.tex
% !TEX spellcheck = en-US

\section{Method}
\label{sec:method}

Given a set of microscopy stacks along with the corresponding and possibly imprecise centerline annotations, we want to train a deep net to produce precise delineation. To this end, when training the deep network, we adjust not only its weights but also the annotations themselves. We first present the vanilla training procedure without annotation adjustment and explain why it is sub-optimal when the annotations lack precision. We then formalize our training procedure with adjustment.

\subsection{Standard Training Procedure}
\label{sec:standard}

Let us consider a set of $N$ microscopy scans $\{\bX_i\}_{1 \leq i \leq N}$ and corresponding centerline annotations $\{\hat\by_i\}_{1 \leq i \leq N}$, in the form of distance maps of the same size as the scans. Voxel $p$ of annotation $\hat\by$, denoted $\hat\by[p]$, contains the distance from the center of $p$ to the closest centerline. Let $F(\cdot;\Theta)$ be a deep network, with weights $\Theta$. It takes a scan $\bX_i$ as input and return a volume $\by_i=F(\bX_i;\Theta)$, containing a delineation of centerlines visible in $\bX_i$. To keep the notation concise, we omit the dependencies on $\by_i$ on $\Theta$. The traditional approach to learning the network weights is to make $\by_i$ as close as possible to $\hat\by_i$ by solving
\begin{equation}
\Theta^*=\argmin_{\Theta}  \sum_{i=1}^N \mathcal{L}\big(\hat\by_i,\by_i\big) , \label{eq:standardObjective}
\end{equation}
where the loss term $\mathcal{L}(\hat\by,\by)$ measures the voxel-wise difference between the annotation and the prediction. In our experiments, we take $\mathcal{L}$ to be the Mean Square Error. 
This assumes that the deviations of the annotations from actual centerline trajectories are small and unbiased. In reality, they rarely are. Hence, the network learns to accommodate this uncertainty in the annotations by blurring the predictions. At test time, this leads to breaking the continuity of predictions wherever the image quality is compromised by high level of noise or low contrast between the foreground and the background, as illustrated by Fig.~\ref{fig:teaser2}. 

\subsection{Overview of our Approach}
\label{sec:proposed}

The formulation of Eq.~\ref{eq:standardObjective} assumes that the deviations of the annotations from reality are small and unbiased. This work is predicated on the fact that they rarely are and that we must allow for substantial non-Gaussian deviations from the original annotations.  Thus, instead of encoding the annotations in terms of volumes $\hat\by_i$, we represent the annotated centerline $\mC_i$ of each $\bX_i$ as a graph, with the set of vertices $\mathcal{V}_i$ and the set of edges $\mathcal{E}_i$. Each vertex $v\in\mathcal{V}_i$ has a 3D coordinate $c_v$, and each edge $(u,v)\in\mathcal{E}_i$ represents a short line segment. This is shown in Fig.~\ref{fig:teaser2} where the circles along the annotations denote the vertices.  Let $\bc_i$ be the vector formed by concatenating coordinates of all the vertices of $\mathcal{V}_i$. To accommodate the possible lack of precision of the annotations, we let $\bc_i$ change its initial value. Doing so changes the shape of $\mC_i$ but preserves its topology and can be used to explicitly model the deviation of the annotated centerlines from their true position. In other words, the minimization problem can be reformulated as finding 
\begin{gather}
\Theta^*,\bC^* = \argmin_{\Theta,\bC}  \sum_{i=1}^N  L(\bc_i,\by_i)+R (\bc_i) , \label{eq:newObjective} \\
\mbox{where } \quad L = \mathcal{L}\big(D(\bc_i),\by_i\big)  \; ; \nonumber \label{eq:ldat}
\end{gather}
$\bC$ is the vector obtained by concatenating all the $\bc_i$; 
$R$ is a regularization term that forces the deformed centerlines to be smooth, and that we define in Sec.~\ref{sec:netsnakes}; 
$\mathcal{L}$ is the same MSE as in Eq.~\ref{eq:standardObjective};
and $D$ is a distance transform that creates a volume in which a voxel with coordinates $q$ is assigned its truncated distance to the closest edge of $\mC$. Formally, we write
\begin{gather}
D(\bc)[q]\!=\!\min\{\delta(\bc,q),d\},  \label{eq:d}\\
\mbox{where}\quad
\delta(\bc,q)\!=\!\min_{(u,v) \in \mathcal{E}} \min_{0\le \phi \le 1}\lVert \phi c_u+(1-\phi) c_v-q\rVert_2 ,
\end{gather}
$d$ is the threshold used to truncate the distance map, and the minimization over $\phi$ serves to find the point on edge $(u,v)$, that is closest to $q$.

Solving the problem of Eq.~\ref{eq:newObjective} means training the network to find centerlines that are smooth and with the same topology as the annotations. This is what we want but, unfortunately, this optimization problem involves two kinds of variables, the components of $\bC$ and $\Theta$ respectively, which are not commensurate in any way. In practice, this makes optimization difficult. We address this problem by eliminating the $\bC$ variables by rewriting Eq.~\ref{eq:newObjective} as
\begin{gather}
\bc^*_i (\by_i) =  \argmin_{\bc} L(\bc_i,\by_i)+R (\bc_i), \label{eq:snakeObjective} \\ 
\Theta^* \!=\!  \argmin_{\Theta}\!  \sum_{i=1}^N \! L\big(\bc^*_i(\by_i),\by_i\big)+R \big(\bc^*_i(\by_i)\big) , \label{eq:ourObjective}
\end{gather}
In the following section, we describe our choice of $R$ and the formulation of $\bc^*_i(\by_i)$ that results from it. Eq.~\ref{eq:ourObjective} is a standard continuous optimization problem that we can solve using the usual tools of the trade. 

\subsection{Annotations as Network Snakes}
\label{sec:netsnakes}

We propose to represent each $\mC_i$ as a network snake, and to take $R$ to be a classical sum of spring and elasticity terms~\cite{Fua96f,Butenuth12}. This regularization term takes the form
\begin{equation} \label{eq:reg}
R(\bc)= \alpha \!\!\sum_{(u,v)\in\mathcal{E}} \!\!\lVert c_u-c_v \rVert^2 + \beta \!\!\sum_{(u,v,w)\in\mathcal{T}} \!\!\lVert c_u-2c_v+c_w \rVert^2,
\end{equation}
where $\alpha$ and $\beta$ are hyper-parameters that balance the strength of the two terms, $\mathcal{E}$ is the set of edges of $\mC$ and $\mathcal{T}$ is the set of node triples $(u,v,w)$ such that $(u,v)\in\mathcal{E}$, $(v,w)\in\mathcal{E}$, and $v$ is a node of order two, that is, not a junction of multiple snake branches.
As shown in~\cite{Fua96f,Butenuth12}, $R$ can be written as
\begin{align} \label{eq:regmatrix}
 R (\bc_i) = \frac{1}{2} \bc_i^T \bA \bc_i ,
 \end{align}
where $A$ is a sparse symmetric matrix. Given this quadratic formulation of $R$, we can use the well-known semi-implicit scheme introduced to deform snakes, also known as active contour models~\cite{Kass88}, to minimize Eq.~\ref{eq:snakeObjective}. It involves initializing each snake $\bc_i^0$ to the manually produced annotation and refining it by iteratively solving
\begin{equation}
(\bA+\gamma \bI) \bc^{t+1}_i=\gamma \bc^t_i - \frac{\partial L}{\partial \bc}(\bc_i^{t},\by_i)  \label{eq:viscosity}
\end{equation}
for $\bc_i^{t+1}$,
where $\gamma$ is a hyper-parameter known as the {\it viscosity} and is inversely proportional to the step size in each iteration.
We refer the reader to~\cite{Kass88} for the complete derivation.
Here we only note, that when the iteration stabilizes, we have $\forall i, \bc^{t}_i \approx \bc^{t+1}_i$.
We can therefore denote the stable vector of node locations by $\bc^*_i$, substitute $\bc^{t+1}_i\approx \bc^{t}_i\approx \bc^{*}_i$ in Eq.~\ref{eq:viscosity}, and use the derivative of Eq.~\ref{eq:regmatrix}, to write
\begin{align}
\forall i, \quad \frac{\partial R}{\partial \bc}(\bc^*_i) + \frac{\partial L}{\partial \bc}(\bc_i^*,\by_i) \approx 0 , \label{eq:zerograd} 
\end{align}
which means that $\bc^*$ minimizes $R+L$ and is a solution of Eq.~\ref{eq:snakeObjective}. 

In practice, we solve Eq.~\ref{eq:viscosity} by inverting the matrix $(\bA+\gamma \bI)$ at the start of the training procedure and then multiplying the right-hand-side of the equation by the inverse at each iteration. Hence, we write
\begin{equation}
\bc^{t+1}_i= 
(\bA+\gamma \bI)^{-1} \big(\gamma\bc^t_i- \frac{\partial L}{\partial \bc} (\bc_i^{t},\by_i)\big). \label{eq:update}
\end{equation}
We perform the update or Eq.~\eqref{eq:update} for $0 \leq t < T$. We take $T=10$ in our implementation, which is sufficient for the process to stabilize, and denote the result of the last iteration by $\bc^*_i(\by_i)=\bc^T_i$.

\subsection{Computing the Gradients of the Loss Function}
\label{sec:grad}

Performing the minimization in Eq.~\ref{eq:ourObjective} requires computing at each iteration the gradient of the loss with respect to the network output $\by_i$.
To avoid cluttering the notation, we denote $\bc^*(\by_i)$ by $\bc^*$. The gradient can then be expressed as
\begin{align}
\frac{\partial}{\partial \by} &\big(L(\bc^*_i,\by_i)+R (\bc^*_i) \big) \nonumber \\
&= \frac{\partial L}{\partial \by}(\bc^*_i,\by_i) + \big(\frac{\partial L}{\partial \bc}(\bc^*_i,\by_i)+\frac{\partial R}{\partial \bc}(\bc^*_i)\big) \frac{\partial \bc^*_i}{\partial \by}   \label{eq:aux1} \\
& \approx \frac{\partial L}{\partial \by}(\bc^*_i,\by_i) , \nonumber 
\end{align}
where we used Eq.~\ref{eq:zerograd} to eliminate the second term. In other words, even though $\bc^*$ is a function of $\by_i$, we do not need to compute its derivatives with respect to $\by_i$ to train the neural network. We only need those of $L$, and can treat $\bc^*$ as a constant when evaluating them. Therefore, the only difference between using our approach and the standard one of Section~\ref{sec:standard} is that instead of evaluating the loss using the original annotation $\bc$, we use its optimized version $\bc^*$. We call this approach \snakeFull{} and it is depicted at the top of Fig.~\ref{fig:grads}.

\input{fig/training}

\subsection{Speeding Things Up}
\label{sec:speedup}

% !TEX root = ../top.tex
% !TEX spellcheck = en-US

\begin{figure}
\center
\setlength{\tabcolsep}{0pt}
\begin{tabular}{@{}cccc@{}}
\multirow{2}{0.24\columnwidth}{\centering \scriptsize Initial dist.\ map and inaccurate annotation} & 
\multicolumn{3}{c}{\scriptsize Snake at convergence and dist.\ map after 100 GD iterations} \\[0mm]
& \scriptsize \snakeFull{} & \scriptsize \snakeFast{} & \scriptsize \snakeSimple{} \\
%\includegraphics[width=0.24\columnwidth]{fig/snake_s_initial} & 
%\includegraphics[width=0.24\columnwidth]{fig/snake_full_s_converged} &
%\includegraphics[width=0.24\columnwidth]{fig/snake_fast_s_converged} &
%\begin{overpic}[width=0.24\columnwidth]{fig/snake_simple_s_converged.png}
%\linethickness{1pt}
%\put(87.5,35){\color{red}\vector(-1,1){10}}
%\end{overpic} 
%\\[3mm]
\includegraphics[width=0.24\columnwidth]{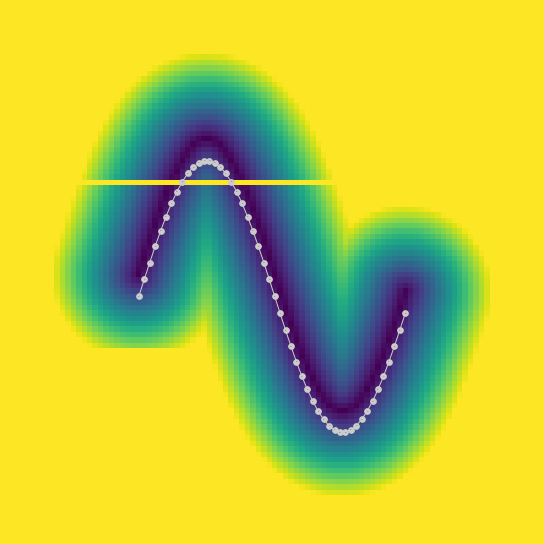} & 
\includegraphics[width=0.24\columnwidth]{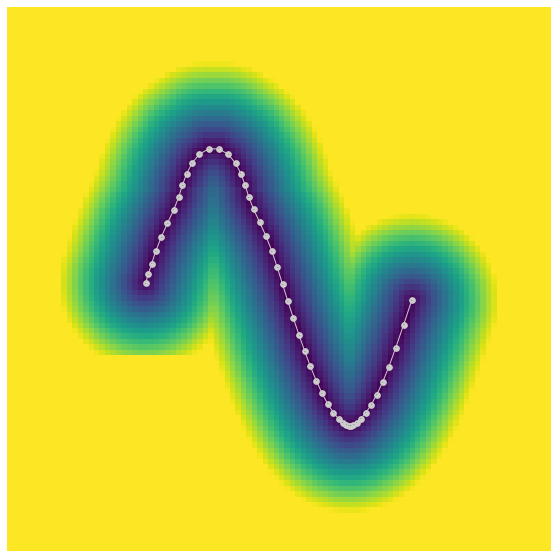} &
\includegraphics[width=0.24\columnwidth]{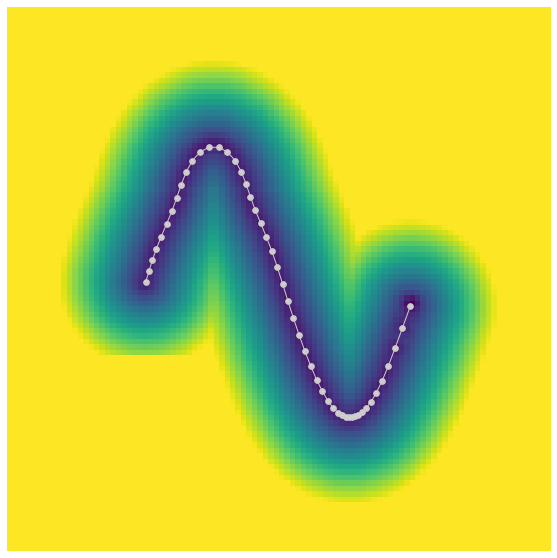} &
\begin{overpic}[width=0.24\columnwidth]{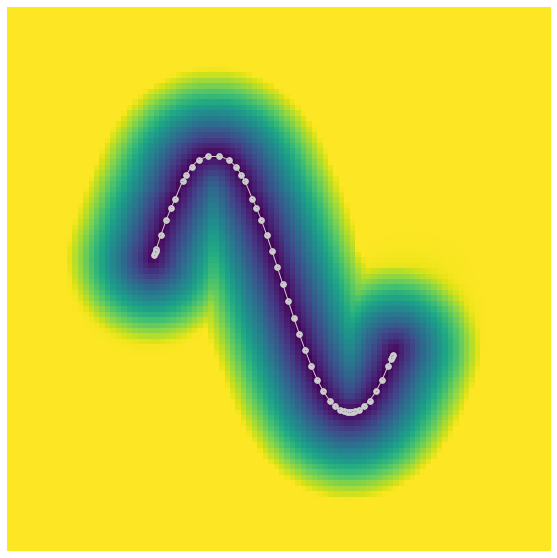}
\linethickness{4pt}
\put(97.5,25){\color{red}\vector(-1,1){20}}
\end{overpic} 
\\[0mm]
\parbox[c][16pt][c]{2.3cm}{\centering \scriptsize GT dist.\ map and accurate annotation} & 
\multicolumn{3}{c}{\scriptsize Difference between the GT dist.\ map and one after 100 GD iters} \\
\includegraphics[width=0.24\columnwidth]{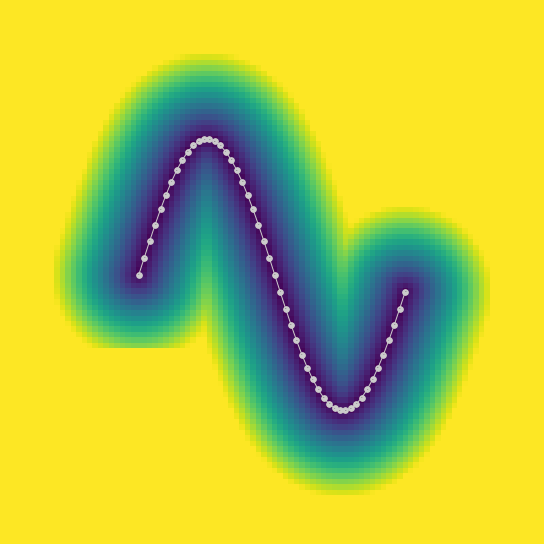} & 
\includegraphics[width=0.24\columnwidth]{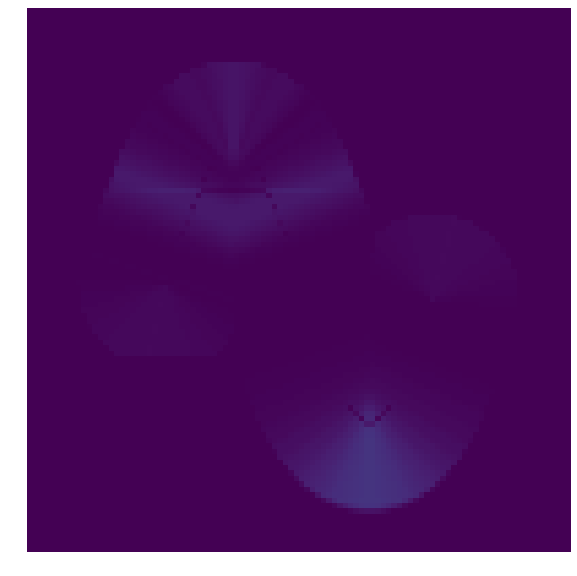} &
\includegraphics[width=0.24\columnwidth]{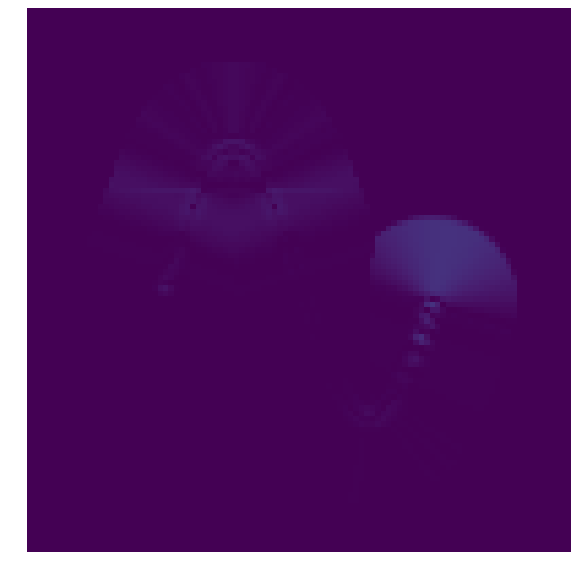} &
\begin{overpic}[width=0.24\columnwidth]{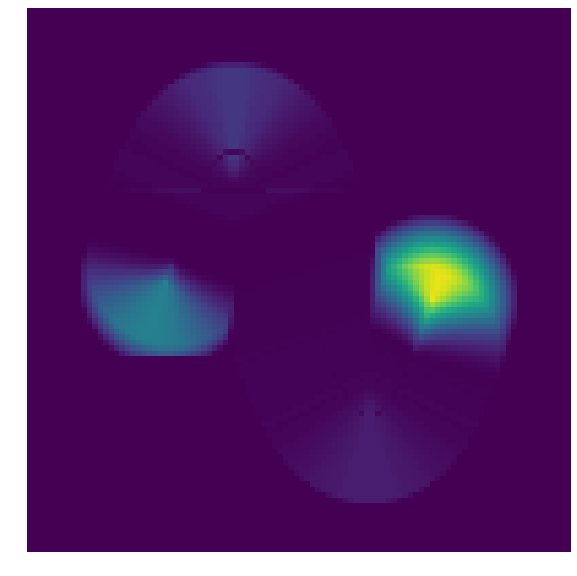}
\linethickness{4pt}
\put(97.5,25){\color{red}\vector(-1,1){20}}
\end{overpic} 
\\[0mm]
%ideal dist.\ map \& annotation 
\multicolumn{1}{r}{\scriptsize computation time [s]:}&
% \parbox{0.08\columnwidth}{time [s]:\hfill} \parbox{0.06\columnwidth}{\hfil 72 \hfil}\parbox{0.08\columnwidth}{\hfill} 
\scriptsize 72
& \scriptsize 12 & \scriptsize 5.3 \\
\end{tabular}
\caption{{\bf Compared behavior of \snakeSimple{}, \snakeFast{}, and \snakeFull{} on a synthetic 2D example.} ({\it Left column}) At the bottom, distance map and corresponding annotation. At the top, we simulated an unwarranted break in the distance map (horizontal yellow line) and shifted the annotation by several pixels. ({\it Other Columns}) In three separate runs, we performed 100 Gradient Descent using either \snakeFull{}, \snakeFast{}, or  \snakeSimple{}. In the {\it top row}, we show the corrected annotation and the updated distance maps. The {\it bottom row} depicts the differences between the updated maps and the ground-truth one. We also indicate the computation times. \snakeFull{} removes the interruption in the distance map but the computation is slow. \snakeFast{} is much faster and fills the gap in the distance map almost as well. \snakeSimple{} is even faster but yields a corrected annotation that is too short, as highlighted by the red arrow. 
}
\label{fig:synt}
\end{figure}

%Columns 2-4 show the modifications of the distance map after 100 updates and the positions to which the proposed snake variants converge in the last iteration. 
%The modifications induced by the loss based on \snakeFull{} are as expected, but as explained in section~\ref{sec:speedup}, the computation of this loss is time consuming.
%The loss based on \snakeFast{} not only fills the gap in the distance map, but also introduces artifacts guiding the snake to a low-energy configuration.
%\snakeSimple{} does not fill the whole valley in the distance map. In consequence, the updates to the distance map tend to shorten the valley.
%In section~\ref{sec:exp} we show that, when the losses are evaluated on real data, the artifacts produced by \snakeFast{} are less detrimental to performance than the ones of \snakeSimple{}, and that \snakeFast{} is substantially faster than \snakeFull{}.

We will show in Section~\ref{sec:exp} that \snakeFull{} performs well but is slow to train. The culprit is the term $\frac{\partial L}{\partial \bc}$ in the update Eq.~\ref{eq:update}, which involves a time-consuming computation of the gradient of a distance map. To speed things up, we introduce a  faster approach that we call \snakeFast{}. In it, we replace the term $L$ in Eq.~\ref{eq:snakeObjective} by 
a simpler objective function $S$ directly inspired by the  classical external snake energy~\cite{Kass88}. We take it to be
\begin{equation}
S(\bc,\by)=\sum_{v \in \mathcal{V}} \big(\by*G\big)[c_v],
\end{equation}
where $*G$ denotes a convolution with a Gaussian kernel and $\by[c_v]$ denotes the network output at vertex $v$. $S$ is very similar to the energies used in traditional network snake formulations~\cite{Fua96f,Butenuth12}. Importantly, $S$ and its gradients are easy and fast to compute because doing so only requires convolving $\by$ with a Gaussian kernel and sampling the result at the locations of the snake nodes.
Deforming the annotations then involves finding 
\begin{align}
\bc^{\dagger}_i(\by_i)          & =  \argmin_{\bc} S (\bc,\by_i) + R (\bc)    ,   \label{eq:snakeFunc} 
\end{align}
which means that the sum of distance values along the snake should be as low as possible while preserving snake smoothness.
%In practice, this yields results that are not strictly equal to those obtained by solving Eq.~\ref{eq:snakeObjective}.
%
As in Section~\ref{sec:netsnakes}, the snake update takes the form
\begin{equation}
\bc^{t+1}_i=(\bA+\gamma \bI)^{-1} \big(\gamma\bc^t_i- \frac{\partial S}{\partial \bc} (\bc_i^{t},\by_i)\big)  . \label{eq:update2}
\end{equation}
In practice, we take $\bc^{\dagger}_i(\by_i)=\bc^T_i$, where $T=10$, as in Section~\ref{sec:netsnakes}. Finally,  we take the network training objective to be  
\begin{align}
\Theta^*                       & =  \argmin_{\Theta} \sum_{i} L (\bc^\dagger_i(\by_i),\by_i) , \label{eq:ourObjective2}
\end{align}
where we still use the original $L$ of Eq.~\ref{eq:newObjective}. We do this because $S$ only depends on a small subset of voxels of $\by$. Hence, it only provides a sparse supervisory signal and is not well suited as the training objective for the network that produces a dense distance map. The gradient of the objective of Eq.~\ref{eq:ourObjective2} is
\begin{equation}
\frac{\partial}{\partial \by} L(\bc^\dagger,\by)  = 
\frac{\partial L}{\partial \by}(\bc^\dagger,\by) + 
\frac{\partial L}{\partial \bc}(\bc^\dagger,\by) \frac{\partial \bc^\dagger}{\partial \by}  . \label{eq:aux2}
\end{equation}
Because we minimized $S$ instead of $L$ in Eq.~\ref{eq:snakeFunc},  we can no longer assume that the second term is zero as we did in Section~\ref{sec:grad}. Hence, to compute it during the minimization,  we backpropagate through the snake update procedure of Eq.~\ref{eq:update2}, as depicted by the middle row of Fig.~\ref{fig:grads}. In practice, we use the autograd functionality of Pytorch to this end.

The non-zero second term of Eq.~\ref{eq:aux2} helps guide the snake to a position where the data loss $L$ is low and ultimately influences the distance map that our deep network $F$ outputs. It could be argued that ignoring this term so that the networks focuses exclusively on fitting the annotations would be preferable. To test this assertion,  we implemented \snakeSimple{}, a third variant or our approach in which we  take the second term of Eq.~\ref{eq:aux2} to be zero. \snakeSimple{} is even faster than \snakeFast{}. In essence, it is a simplified version of \snakeFull{} and \snakeFast{} in which we successively optimize the network weights and then the snake position without any direct interaction between these two optimization steps. 

Fig.~\ref{fig:synt} uses a synthetic example to illustrates the differences between our three variants. \snakeFast{} and \snakeFull{} yield similar results with the former being much faster whereas \snakeSimple{} is even faster but prone to generating artifacts. We now turn to our experimental results on real data that confirm this.

%To better expose the differences between the three snake variants, we evaluated the three associated loss functions on a synthetic pair of a perturbed distance map with a misaligned annotation. 
%To minimize the loss function we applied 100 gradient descent iterations to the distance map.
%As expected, \snakeFull{} acts to fix the interruption in the distance map.
%The distance map produced by \snakeFast{} shows artifacts that follow from optimizing the distance map not only for compatibility with the annotation, but also to guide the snake to the position compatible with the distance map.  \snakeSimple{} is free from such artifacts, but the snake converges to a position that does not match the distance map perfectly. Over many update iterations, this results in unwelcome changes to the distance map.

% !TEX root = ../top.tex
% !TEX spellcheck = en-US

\section{Experiments}
\label{sec:exp}

\subsection{Datasets}
\label{sec:datasets}

We tested our approach on the following data sets.
\begin{itemize}

 \item The \neurons{} data set comprises fourteen two-photon microscopy 3D scans of fragments of a mouse brain, with manually traced neurites. We use four volumes for testing and ten for training, each of size $200\times250\times250$ voxels and spatial resolution $0.3\times0.3\times1.0~\mu m$. 

 \item The \neuronsmisal{} data set contains two 3D images of neurons in a mouse brain. They had been outlined manually while viewing the sample under a microscope and the image was captured later. The sample deformed in the meantime, exacerbating misalignment between the annotation and the image. We use one stack of size $151\times714\times865$ voxels and a resolution of $1~\mu m$ for training and one of size $228\times764\times1360$ for testing. 
 
\item The \mra{} is a publicly available set of Magnetic Resonance Angiography brain scans~\cite{Bullitt05}. It consists of 42 annotated stacks, which we cropped to $416 \times 320 \times 128$ voxels by removing their empty margins. Their resolution is $0.5 \times 0.5 \times 0.6$~mm. We randomly partitioned the data into 31 training and 11 test volumes. 

%\item The \synth{} data set contains thirty 3D synthetically generated vasculature volumes. The data is produced by using {\it VascuSynth} tree generator~\cite{Hamarneh10}. We use twenty stacks for training and ten for testing, each of size 400x400x400. Fig.~\ref{fig:visual_synth} shows the maximum-intensity projection of a test stack.

% \item \bcarl{}.  It contains three 3D images of three different mouse brains. The axons and dendrites are manually annotated. We trained our network with 30 volumes obtained from two brain images and we used 10 volumes from the other image for testing. All the volumes are of size of $250\times250\times250~pixels$.

\end{itemize}

None of our data sets can be considered as perfectly annotated. All annotations were performed as accurately as possible, but their precision is affected by the uneven distribution of the dye, image noise, and generic difficulty of annotating 3D volumes. In \neuronsmisal{}, the difficulty is compounded by the fact that the annotation were performed live days before image acquisition, and the sample deformed in the meantime. 

\subsection{Metrics}
\label{sec:metrics}

We used the following performance metrics.
\begin{itemize}
\item \CCQ{}.  
Since standard segmentation metrics such as the F1 score~\cite{Seyedhosseini13} and precision-recall break-even point~\cite{Mnih13} are very sensitive to misalignment of thin structures, we use the {\it correctness-completeness-quality}, which is specifically designed for linear structures~\cite{Wiedemann98}. Correctness corresponds to precision, completeness to recall, and quality to the intersection-over-union. However, the notion of a true positive is relaxed from perfect coincidence of the ground truth and the prediction to their co-occurrence within a distance of $d$ pixels. We used $d=3$. Although it accounts for possible ground truth misalignment, \CCQ{} is still a voxel-wise metric, insensitive to topological errors, such as short interruptions of neurites.
\item \APLS{}. 
The {\it Average Path Length Similarity} is defined as the aggregation of relative length differences of shortest paths between pairs of corresponding end points, randomly sampled in the reconstructed and predicted graphs. It was introduced to evaluate road map reconstructions from aerial images~\cite{APLS} and aims to evaluate the connectivity of the reconstructions, as opposed to their pixel-wise accuracy, which makes it a perfect performance measure for our task.
\item \TLTS{}.
The {\it Too-Long-Too-Short} is another performance criterion based on statistics of relative lengths of shortest paths between corresponding pairs of end points in the prediction and the ground truth~\cite{Wegner13}. 
We report the fraction of {\it correct} paths, that is, predicted paths whose relative length difference to the corresponding ground truth paths is lower than 15\%.
%The predicted paths are classified with respect to their length, relative to the corresponding ground truth paths, as too long, too short, correct, or infeasible, if a ground truth path is not reflected in the prediction. The single-number value we use to evaluate our results is the fraction of {\it correct} paths, with relative length difference not larger than 15\%. 
\end{itemize}

\subsection{Architectures and Training Details}

% !TEX root = ../top.tex
% !TEX spellcheck = en-US

\begin{table}
\caption{
Performance of deep nets trained with different loss functions on our three data sets and the time needed for single training iteration.
 \label{tab:results_combined}
}
\setlength{\tabcolsep}{3pt}
\center
\begin{tabular}{@{}l l@{}l  c c c  c c c  c  c c @{}}
%\cmidrule{4-11}
&  & & \multicolumn{3}{c}{Pixel-wise} & &  \multicolumn{2}{c}{Topology-aware} & & iter.\ t.\\
\cmidrule{4-6}
\cmidrule{8-9}
\cmidrule{11-11}
& \multicolumn{2}{c}{Method} & Corr. & Compl.  & Qual.&   &APLS & TLTS & & s\\
\cmidrule{1-11}

%\parbox[t]{2mm}{
  \multirow{11}{*}{\rotatebox[origin=rb]{90}{\neurons{}}}
%}

&\UNet{}--&\baseline{}        & 98.9 & 91.3 &     90.4 &&     80.3 &     80.9 && 2.8 \\ 
&\UNet{}--&\snakeSimple{}     & 98.4 & 92.5 &     91.2 &&     84.2 &     83.4 && 3.8 \\
&\UNet{}--&\snakeFull{}       & 99.0 & 94.4 &     93.5 &&     89.3 &     85.9 &&18.9 \\
&\UNet{}--&\snakeFast{}       & 98.7 & 95.0 &     93.8 &&     91.1 &     85.9 && 5.2 \\
%\cmidrule{2-11}
\rule{0pt}{3ex}
&\DRU{}--&\baseline{}         & 97.2 & 94.0 &     91.5 &&     84.3 &     83.9 && 2.7 \\ 
&\DRU{}--&\snakeSimple{}      & 97.4 & 95.2 &     92.9 &&     90.8 &     85.9 && 3.8 \\
&\DRU{}--&\snakeFull{}        & 96.9 & 96.9 &     94.1 &&{\bf 91.8}&{\bf 89.3}&&19.1 \\
&\DRU{}--&\snakeFast{}        & 97.0 & 97.1 &     94.2 &&     91.7 &     88.1 && 5.3 \\
%\cmidrule{2-11}
\rule{0pt}{3ex}
&\multicolumn{2}{l}{{\COPLE{}}} & {97.7} & {97.0} &{{\bf 94.8}}&&     {81.0} &     {83.6} && {2.8} \\ 
&\multicolumn{2}{l}{{\QAM{}}  } & {94.5} & {98.8} &     {93.5} &&     {87.3} &     {84.5} && {4.2} \\ 
&\multicolumn{2}{l}{{\DS{}}   } & {97.5} & {97.0} &     {94.7} &&     {83.8} &     {84.1} && {5.8} \\ 
\cmidrule{1-11}

%\parbox[t]{2mm}{
  \multirow{11}{*}{\rotatebox[origin=rb]{90}{\neuronsmisal{}}}
%}

&\UNet{}--&\baseline{}        & 81.8 & 83.5 &     70.4 &&     65.8 &     63.6 && 2.8 \\ 
&\UNet{}--&\snakeSimple{}     & 83.0 & 83.9 &     71.6 &&     70.4 &     68.8 && 3.4 \\
&\UNet{}--&\snakeFull{}       & 83.5 & 85.4 &     73.1 &&     74.2 &     69.9 &&17.8 \\
&\UNet{}--&\snakeFast{}       & 83.1 & 85.5 &     72.9 &&     73.9 &     70.2 && 4.9 \\
%\cmidrule{2-10}
\rule{0pt}{3ex}
&\DRU{}--&\baseline{}         & 82.1 & 86.5 &     72.8 &&     68.9 &     69.5 && 2.7 \\ 
&\DRU{}--&\snakeSimple{}      & 83.2 & 87.7 &     74.5 &&     73.8 &     74.6 && 3.5 \\
&\DRU{}--&\snakeFull{}        & 84.4 & 88.5 &     76.1 &&     74.8 &{\bf 78.1}&&18.3 \\
&\DRU{}--&\snakeFast{}        & 84.2 & 88.9 &{\bf 76.2}&&{\bf 75.1}&     77.7 && 5.1 \\
%\cmidrule{2-10}
\rule{0pt}{3ex}
&\multicolumn{2}{l}{{\COPLE{}}} & {85.2} & {83.4} &     {72.8} &&     {67.6} &     {65.2} && {2.8} \\ 
&\multicolumn{2}{l}{{\QAM{}}  } & {89.8} & {79.3} &     {72.8} &&     {71.2} &     {68.6} && {4.2} \\ 
&\multicolumn{2}{l}{{\DS{}}   } & {83.2} & {81.6} &     {70.0} &&     {71.0} &     {68.8} && {5.8} \\ 
\cmidrule{1-11}

%\parbox[t]{2mm}{
  \multirow{11}{*}{\rotatebox[origin=rb]{90}{\mra{}}}
%}

&\UNet{}--&\baseline{}        & 90.1 & 72.2 &     66.9 &&     49.8 &     50.4 && 2.8 \\ 
&\UNet{}--&\snakeSimple{}     & 89.9 & 73.1 &     67.5 &&     53.5 &     53.1 && 3.7 \\
&\UNet{}--&\snakeFull{}       & 90.2 & 73.5 &     68.0 &&{\bf 55.6}&     55.0 &&18.5 \\
&\UNet{}--&\snakeFast{}       & 90.3 & 73.5 &     68.1 &&     55.4 &     55.2 && 5.1 \\
%\cmidrule{2-10}
\rule{0pt}{3ex}
&\DRU{}--&\baseline{}         & 80.2 & 79.3 &     66.3 &&     48.7 &     49.9 && 2.7 \\ 
&\DRU{}--&\snakeSimple{}      & 80.7 & 79.9 &     67.1 &&     53.3 &     53.0 && 3.7 \\
&\DRU{}--&\snakeFull{}        & 80.9 & 80.5 &     67.6 &&{\bf 55.6}&     55.2 &&18.8 \\
&\DRU{}--&\snakeFast{}        & 81.0 & 80.5 &     67.7 &&     55.3 &{\bf 55.4}&& 5.2 \\

\rule{0pt}{3ex}
&\multicolumn{2}{l}{{\COPLE{}}} & {85.5} & {77.3} &{{\bf 68.3}}&&     {50.2} &     {53.8} && {2.8} \\ 
&\multicolumn{2}{l}{{\QAM{}}  } & {80.2} & {80.1} &     {66.8} &&     {54.3} &     {54.2} && {4.2} \\ 
&\multicolumn{2}{l}{{\DS{}}   } & {82.0} & {80.3} &     {68.1} &&     {55.0} &     {54.9} && {5.8} \\ 
\cmidrule{1-11}

\end{tabular}
\end{table}

% !TEX root = ../top.tex
% !TEX spellcheck = en-US

\begin{figure*}[!htb]
\centering
\begin{tabular}{@{}>{\centering\arraybackslash}m{0.02\textwidth} 
                @{}>{\centering\arraybackslash}m{0.1625\textwidth} 
                @{}>{\centering\arraybackslash}m{0.1625\textwidth}
                @{}>{\centering\arraybackslash}m{0.1625\textwidth} 
                @{}>{\centering\arraybackslash}m{0.1625\textwidth} 
                @{}>{\centering\arraybackslash}m{0.1625\textwidth}
                @{}>{\centering\arraybackslash}m{0.1625\textwidth} @{}}
& \multicolumn{2}{c}{\neurons{}} & \multicolumn{2}{c}{\neuronsmisal{}} & \multicolumn{2}{c}{\mra{}} \\

\rotatebox[origin=l]{90}{\footnotesize \emph{input + annot.}} &
\includegraphics[height=0.1625\textwidth,angle=90]{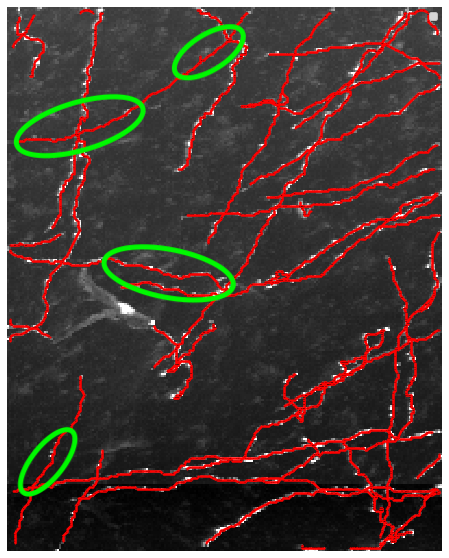} &
\includegraphics[height=0.1625\textwidth,angle=90]{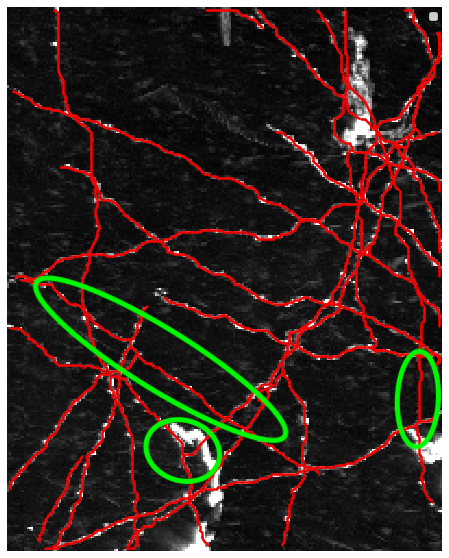} &
\includegraphics[height=0.1625\textwidth,angle=90]{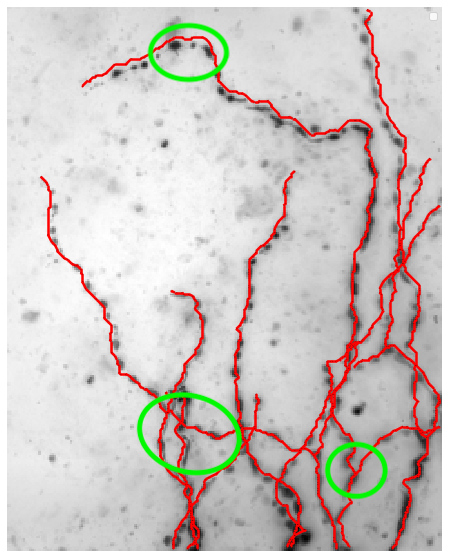} &
\includegraphics[height=0.1625\textwidth,angle=90]{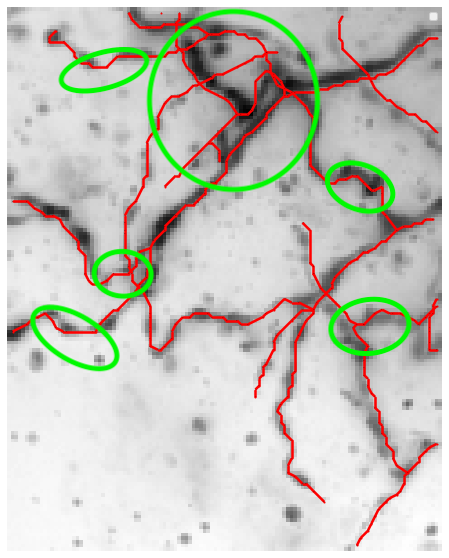} &
\includegraphics[height=0.1625\textwidth,angle=90]{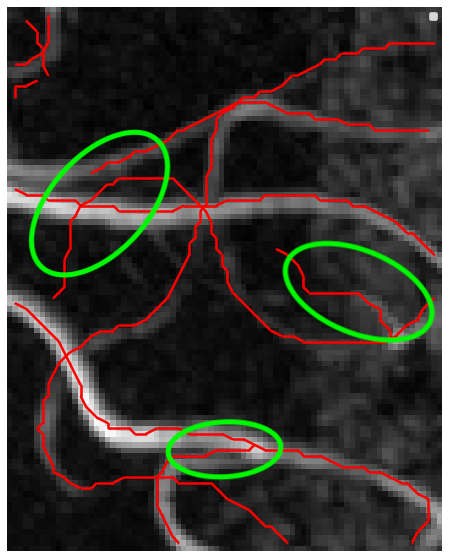} &
\includegraphics[height=0.1625\textwidth,angle=90]{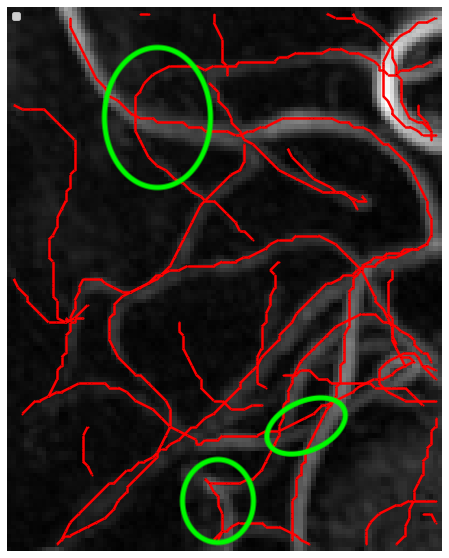} \\

{\rotatebox[origin=l]{90}{\footnotesize \UNet{}+\baseline{} }} &
\includegraphics[height=0.1625\textwidth,angle=90]{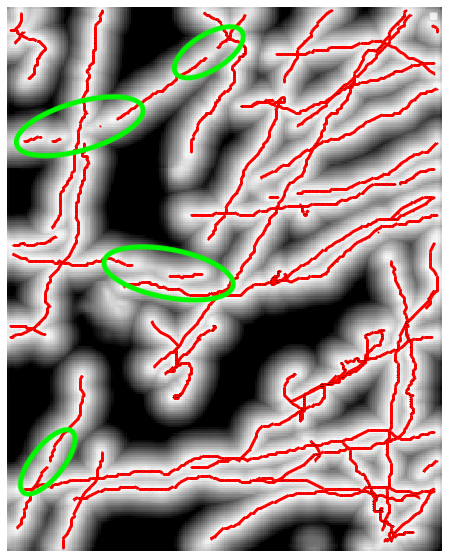} &
\includegraphics[height=0.1625\textwidth,angle=90]{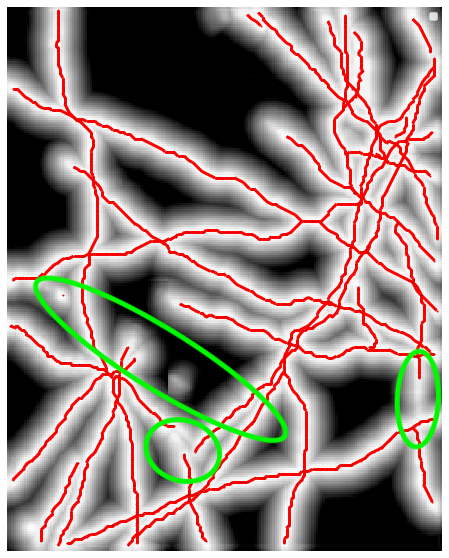} &
\includegraphics[height=0.1625\textwidth,angle=90]{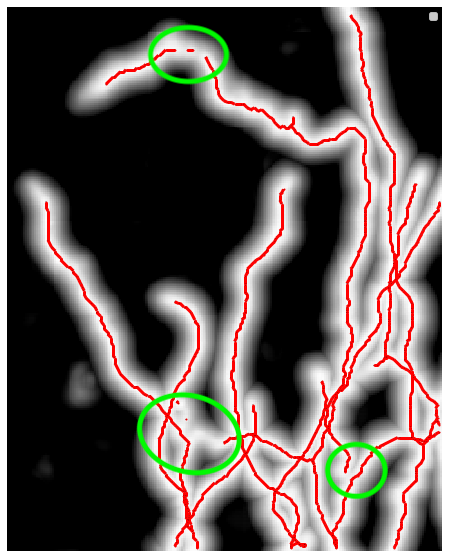} &
\includegraphics[height=0.1625\textwidth,angle=90]{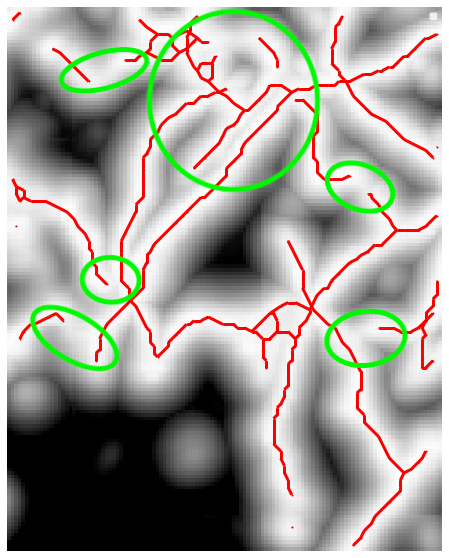} &
\includegraphics[height=0.1625\textwidth,angle=90]{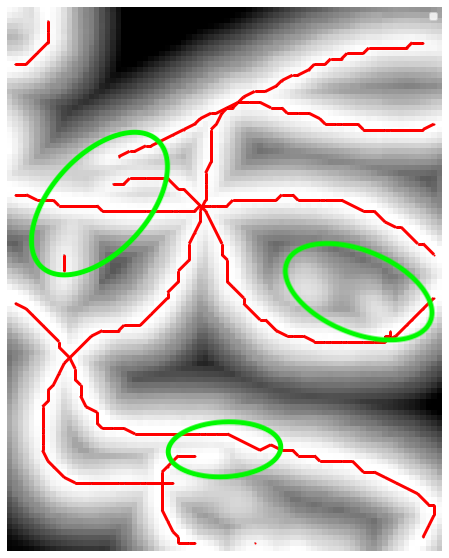} &
\includegraphics[height=0.1625\textwidth,angle=90]{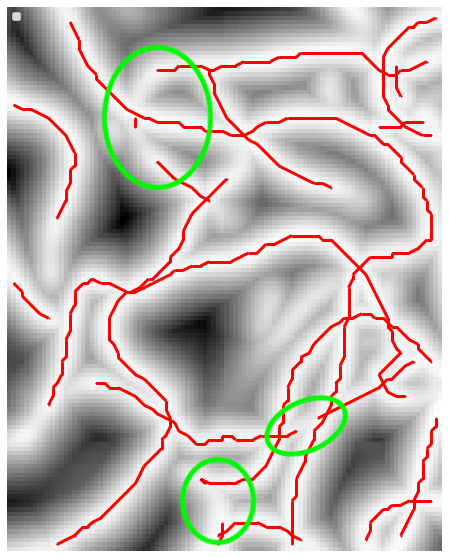} \\

\rotatebox[origin=l]{90}{\footnotesize \UNet{}+\snakeSimple{} } &
\includegraphics[height=0.1625\textwidth,angle=90]{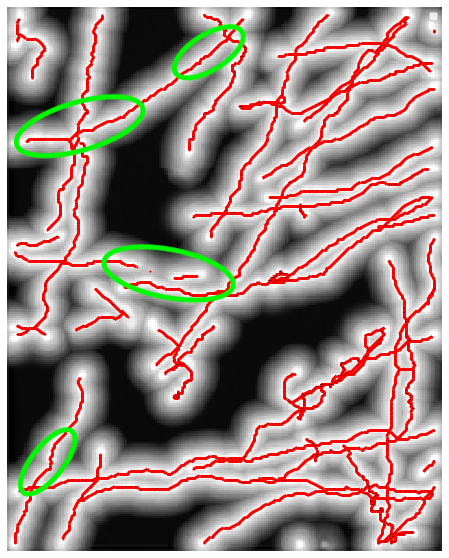} &
\includegraphics[height=0.1625\textwidth,angle=90]{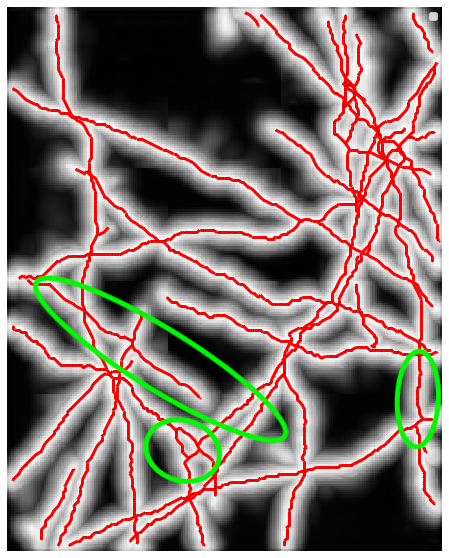} &
\includegraphics[height=0.1625\textwidth,angle=90]{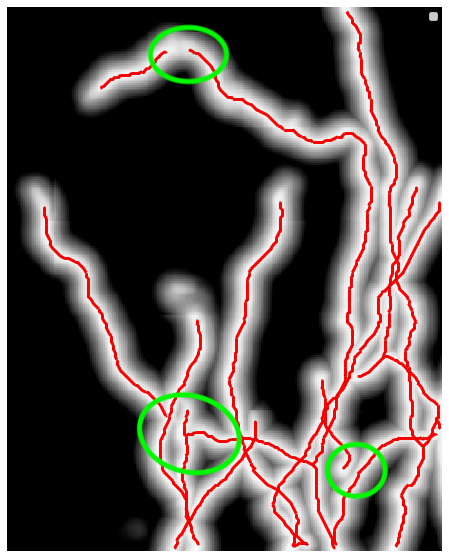} &
\includegraphics[height=0.1625\textwidth,angle=90]{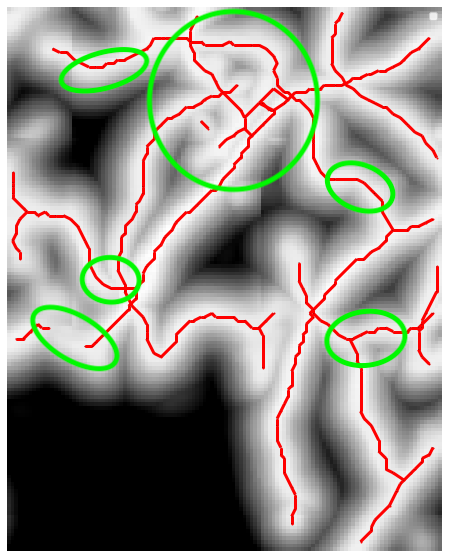} &
\includegraphics[height=0.1625\textwidth,angle=90]{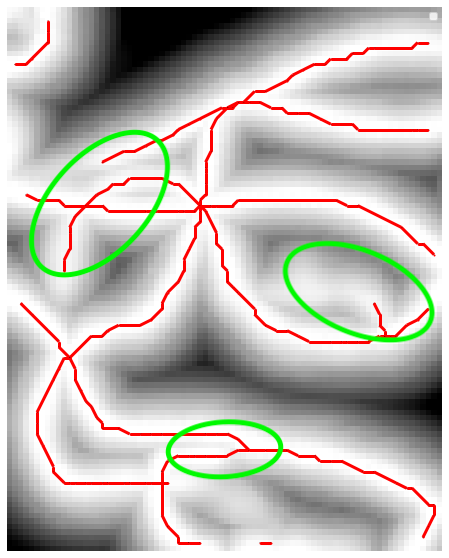} &
\includegraphics[height=0.1625\textwidth,angle=90]{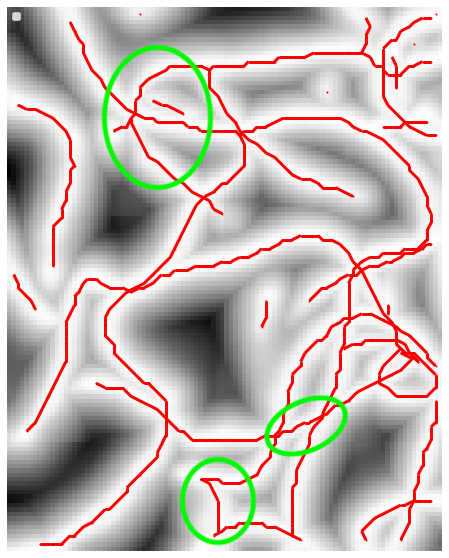} \\

\rotatebox[origin=l]{90}{\footnotesize \UNet{}+\snakeFull{} } &
\includegraphics[height=0.1625\textwidth,angle=90]{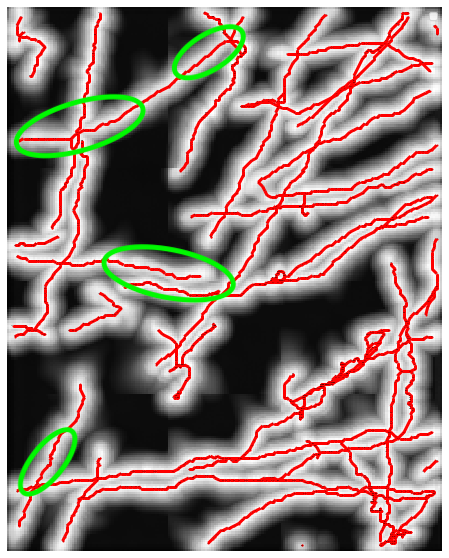} &
\includegraphics[height=0.1625\textwidth,angle=90]{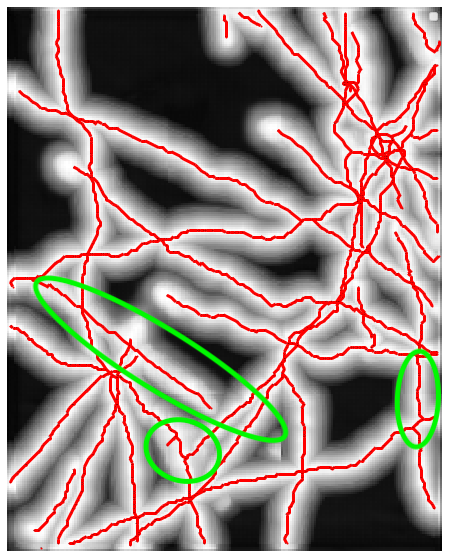} &
\includegraphics[height=0.1625\textwidth,angle=90]{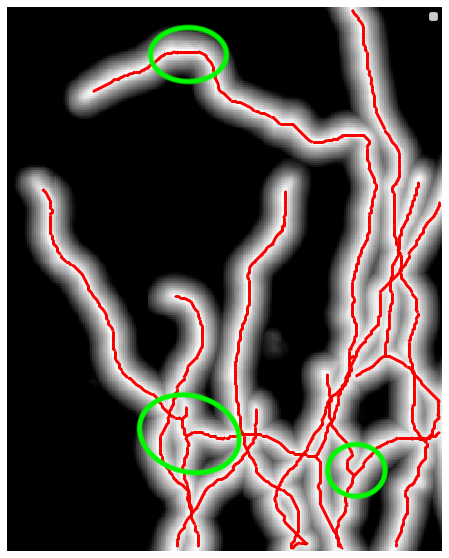} &
\includegraphics[height=0.1625\textwidth,angle=90]{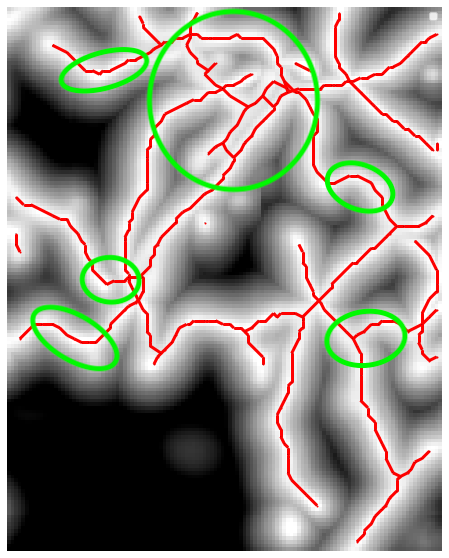} &
\includegraphics[height=0.1625\textwidth,angle=90]{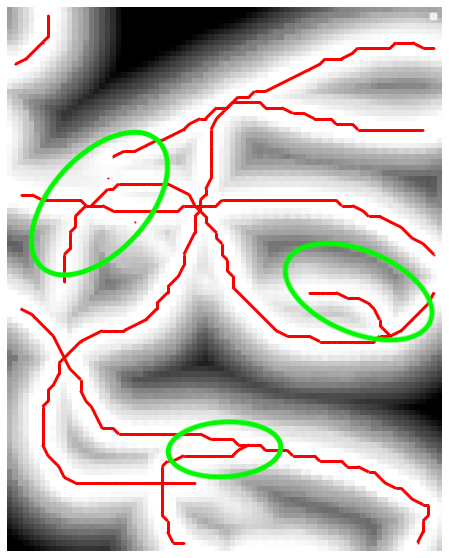} &
\includegraphics[height=0.1625\textwidth,angle=90]{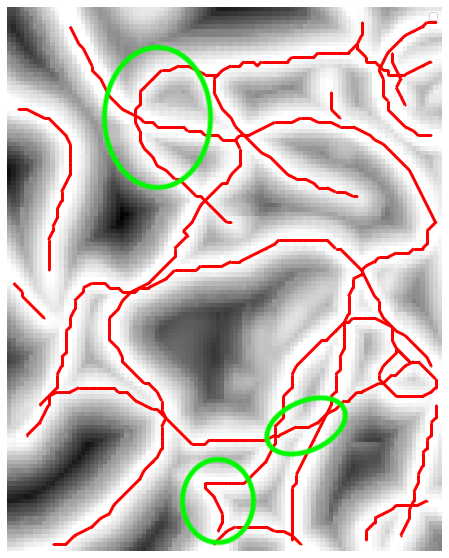} \\

\rotatebox[origin=l]{90}{\footnotesize \UNet{}+\snakeFast{} } &
\includegraphics[height=0.1625\textwidth,angle=90]{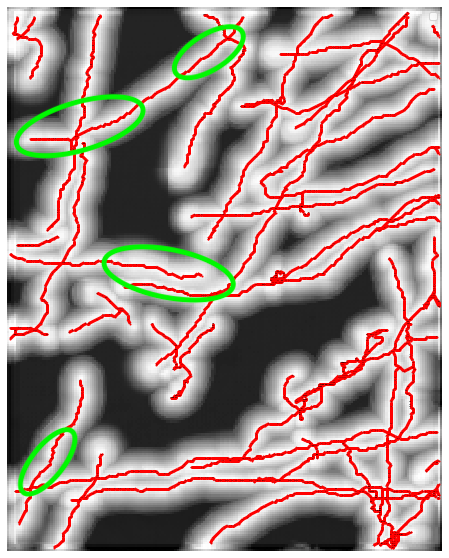} &
\includegraphics[height=0.1625\textwidth,angle=90]{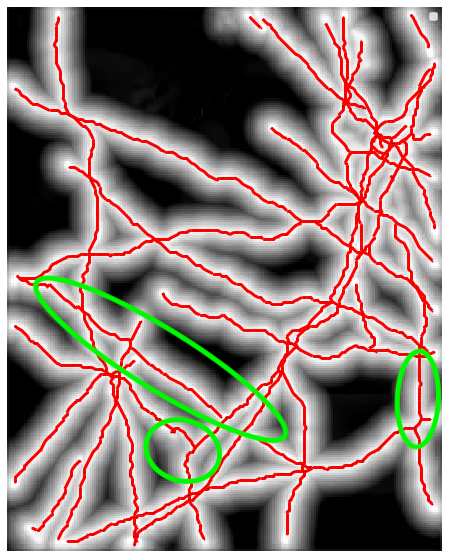} &
\includegraphics[height=0.1625\textwidth,angle=90]{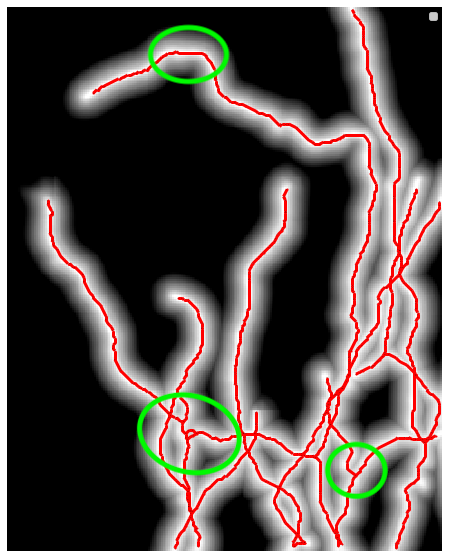} &
\includegraphics[height=0.1625\textwidth,angle=90]{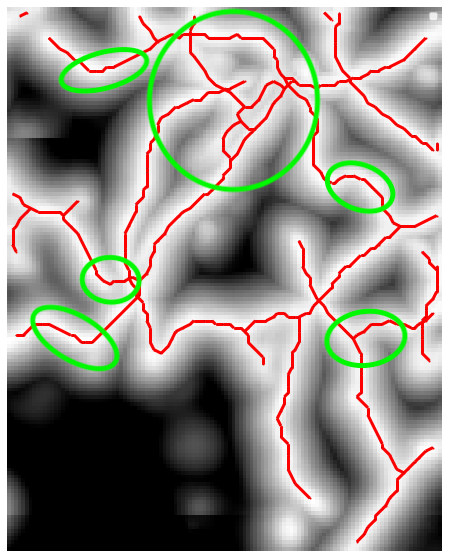} &
\includegraphics[height=0.1625\textwidth,angle=90]{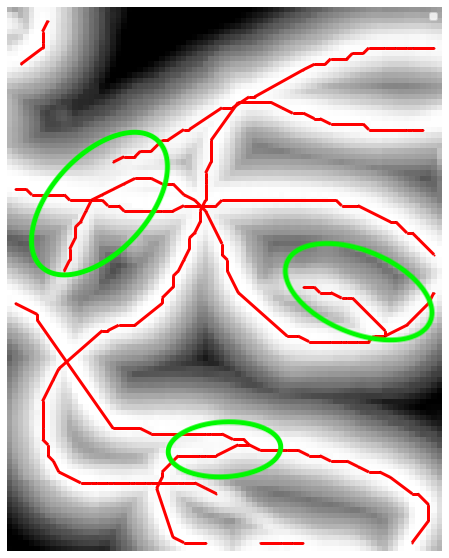} &
\includegraphics[height=0.1625\textwidth,angle=90]{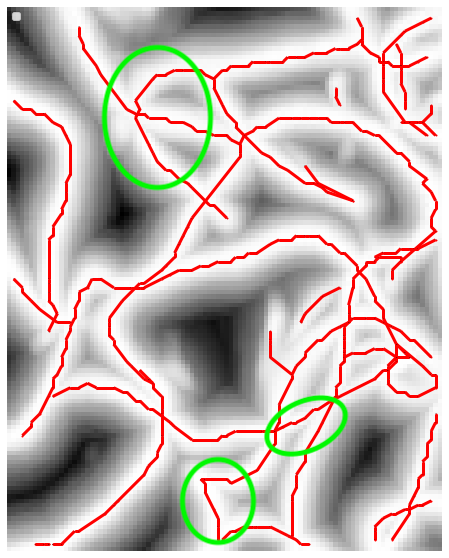} \\

{\rotatebox[origin=l]{90}{\footnotesize \DRU{}+\baseline{} }} &
\includegraphics[height=0.1625\textwidth,angle=90]{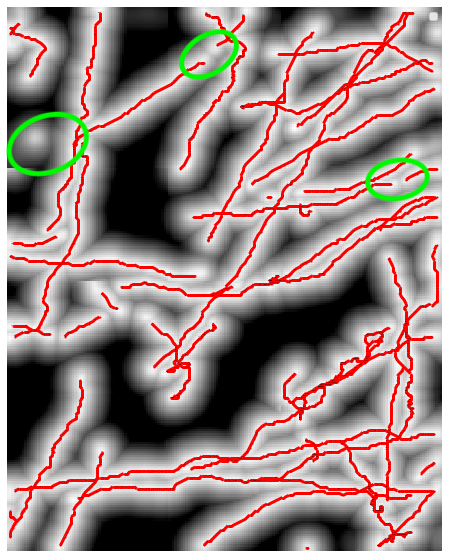} &
\includegraphics[height=0.1625\textwidth,angle=90]{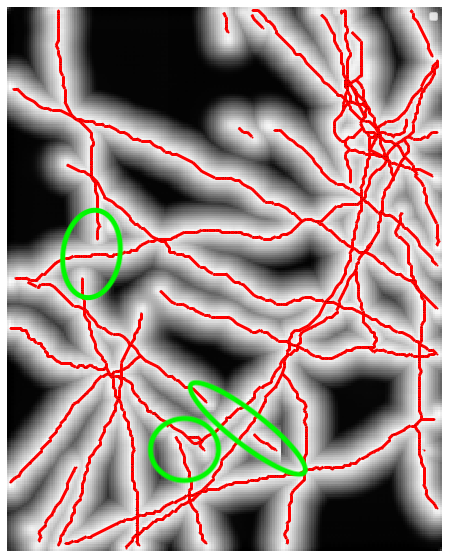} &
\includegraphics[height=0.1625\textwidth,angle=90]{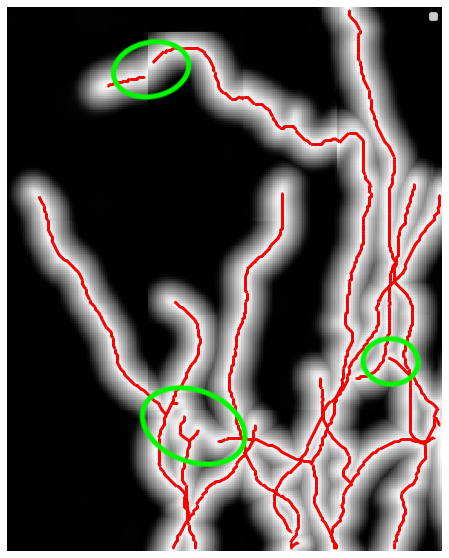} &
\includegraphics[height=0.1625\textwidth,angle=90]{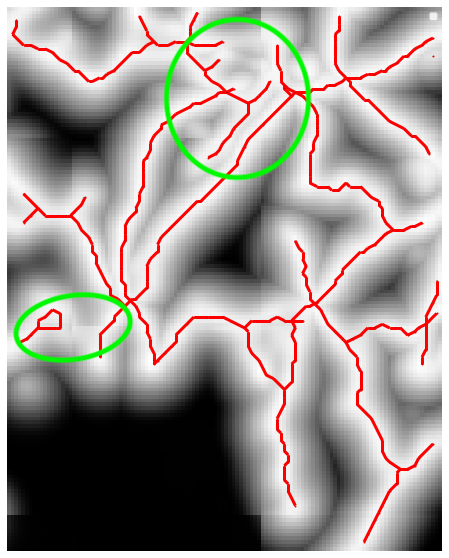} &
\includegraphics[height=0.1625\textwidth,angle=90]{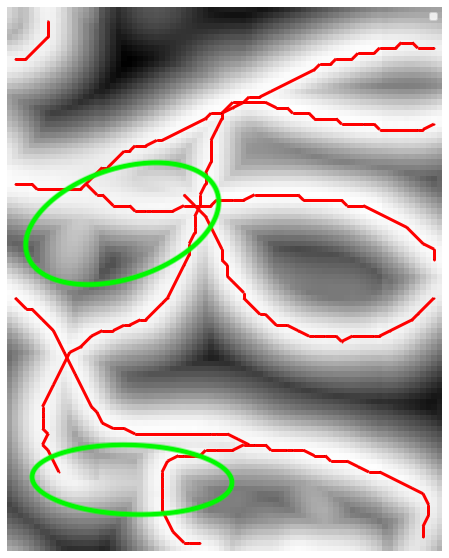} &
\includegraphics[height=0.1625\textwidth,angle=90]{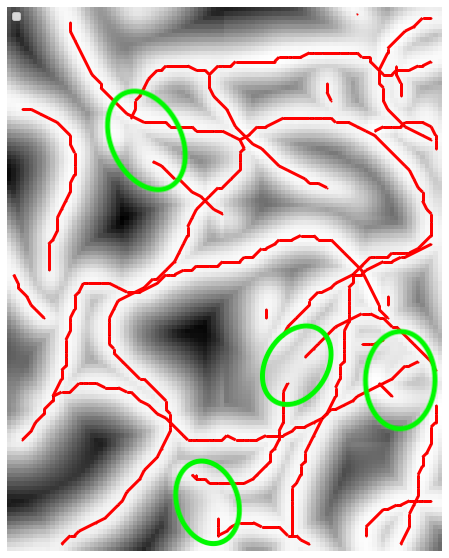} \\

\rotatebox[origin=l]{90}{\footnotesize \DRU{}+\snakeSimple{} } &
\includegraphics[height=0.1625\textwidth,angle=90]{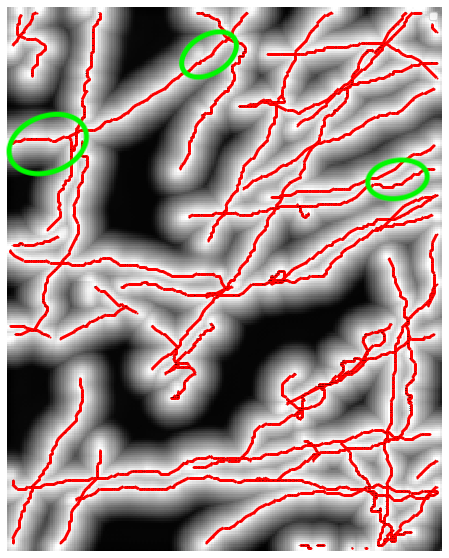} &
\includegraphics[height=0.1625\textwidth,angle=90]{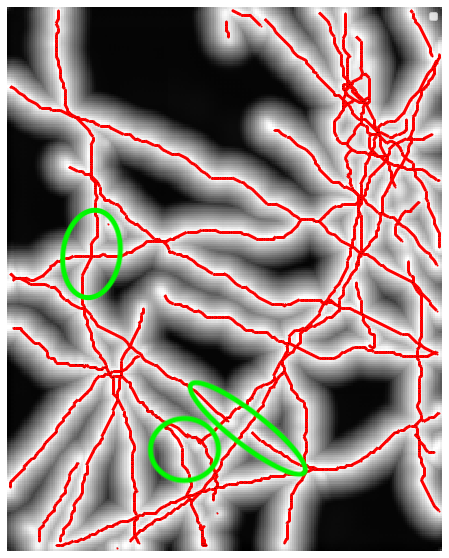} &
\includegraphics[height=0.1625\textwidth,angle=90]{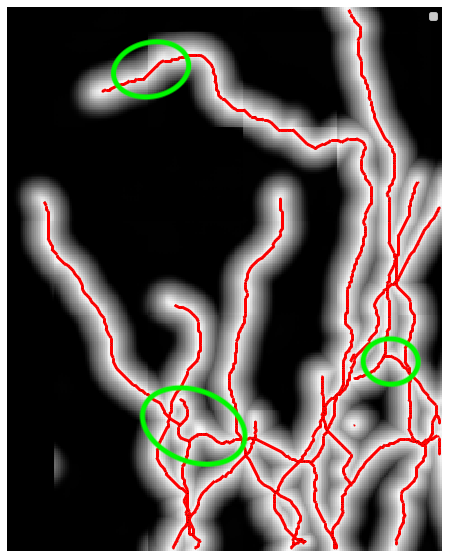} &
\includegraphics[height=0.1625\textwidth,angle=90]{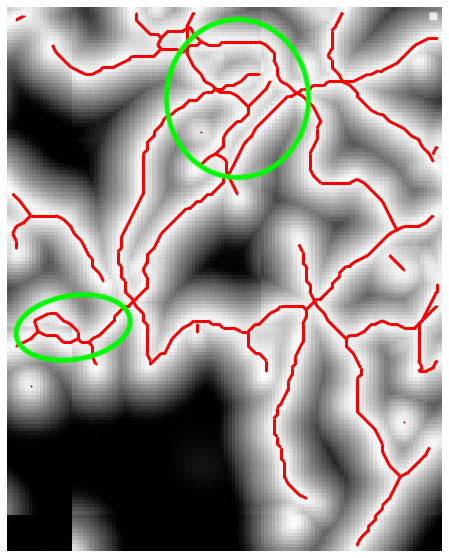} &
\includegraphics[height=0.1625\textwidth,angle=90]{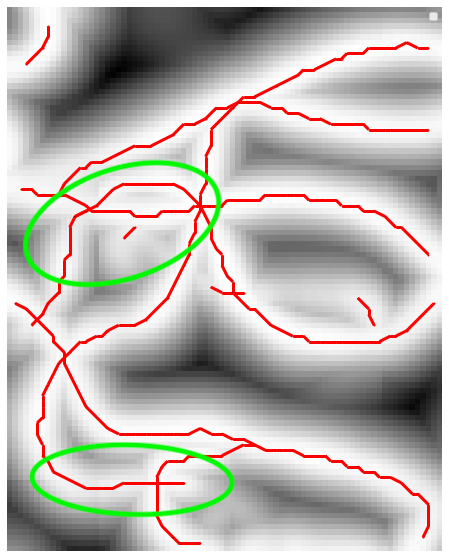} &
\includegraphics[height=0.1625\textwidth,angle=90]{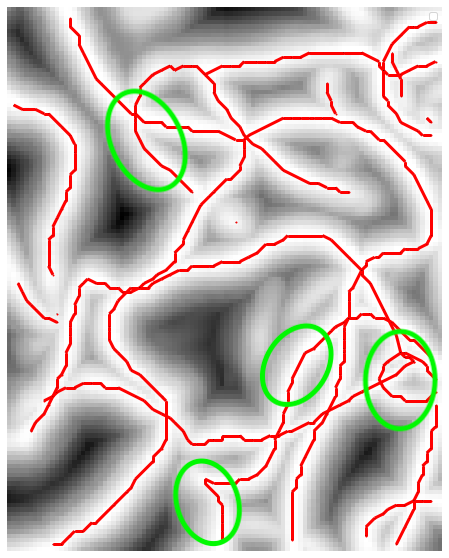} \\

\rotatebox[origin=l]{90}{\footnotesize \DRU{}+\snakeFull{} } &
\includegraphics[height=0.1625\textwidth,angle=90]{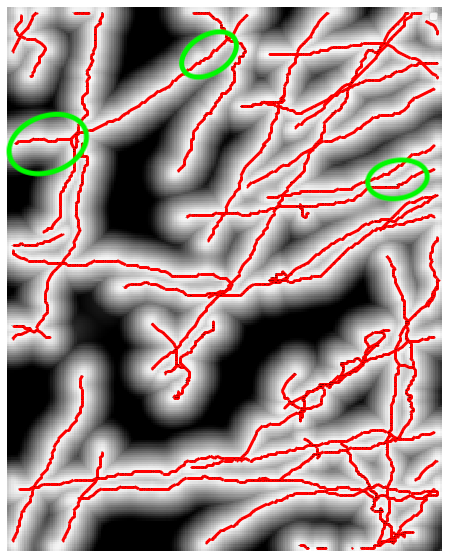} &
\includegraphics[height=0.1625\textwidth,angle=90]{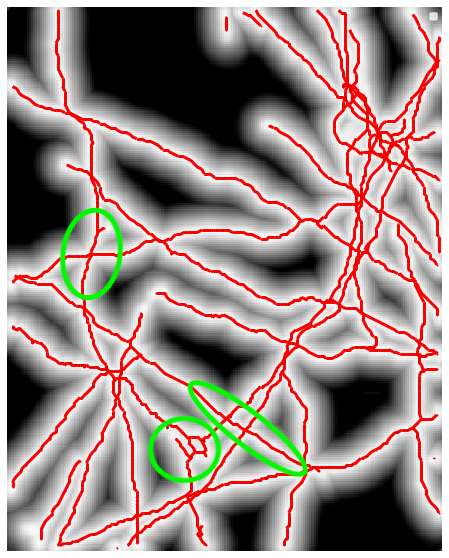} &
\includegraphics[height=0.1625\textwidth,angle=90]{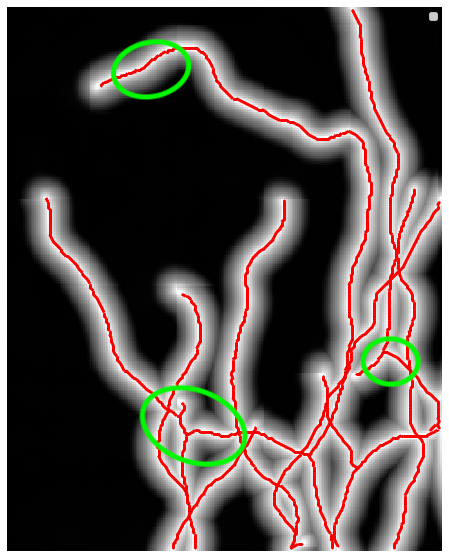} &
\includegraphics[height=0.1625\textwidth,angle=90]{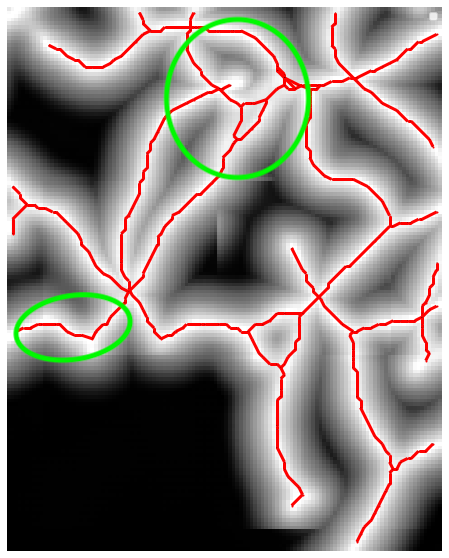} &
\includegraphics[height=0.1625\textwidth,angle=90]{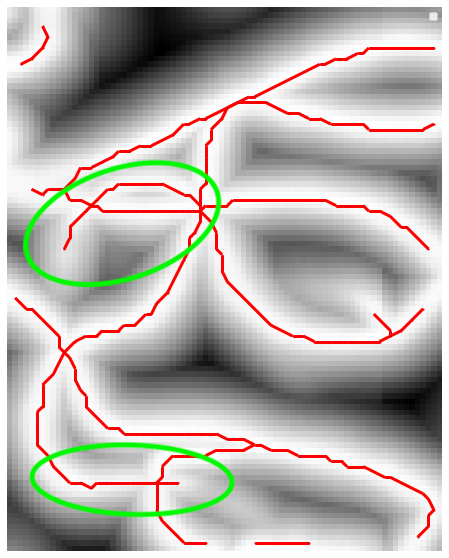} &
\includegraphics[height=0.1625\textwidth,angle=90]{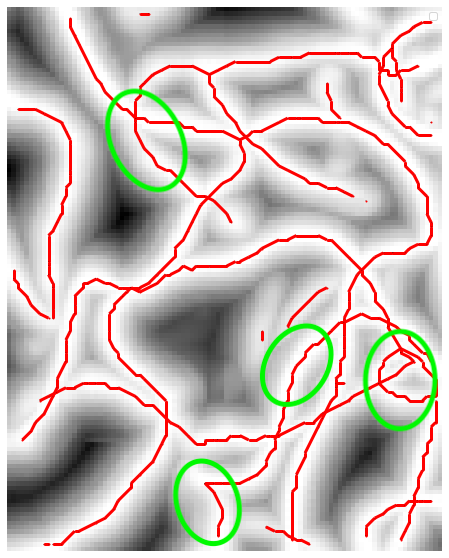} \\

\rotatebox[origin=l]{90}{\footnotesize \DRU{}+\snakeFast{} } &
\includegraphics[height=0.1625\textwidth,angle=90]{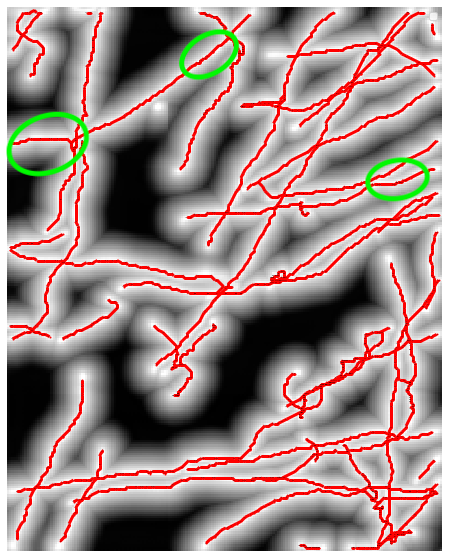} &
\includegraphics[height=0.1625\textwidth,angle=90]{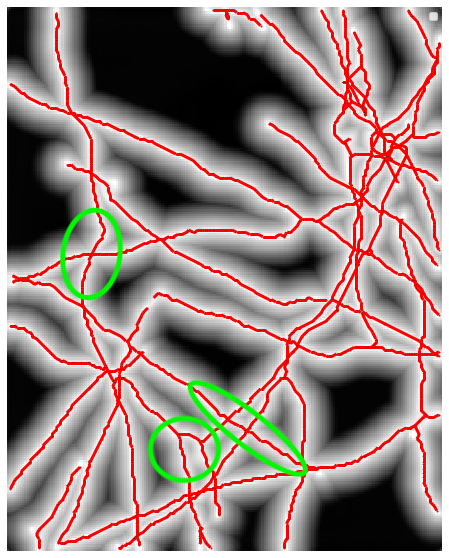} &
\includegraphics[height=0.1625\textwidth,angle=90]{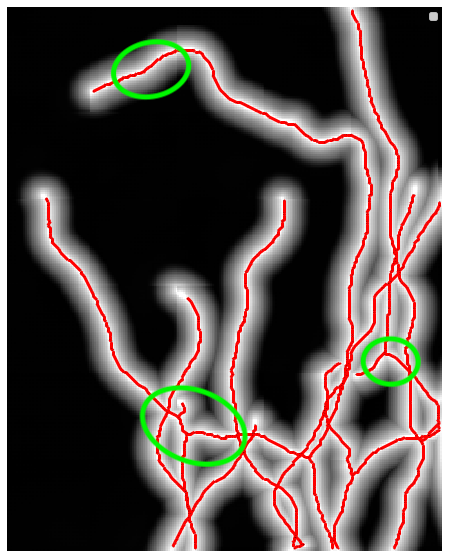} &
\includegraphics[height=0.1625\textwidth,angle=90]{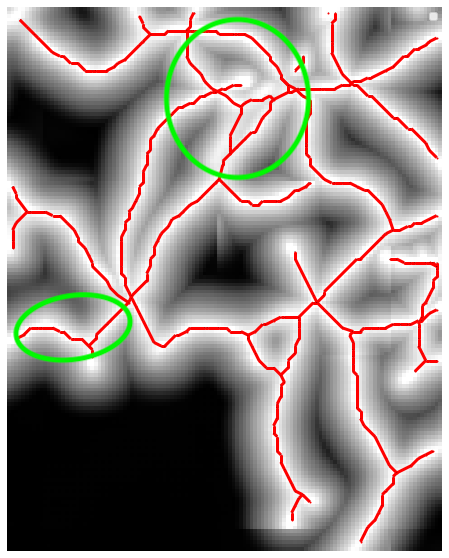} &
\includegraphics[height=0.1625\textwidth,angle=90]{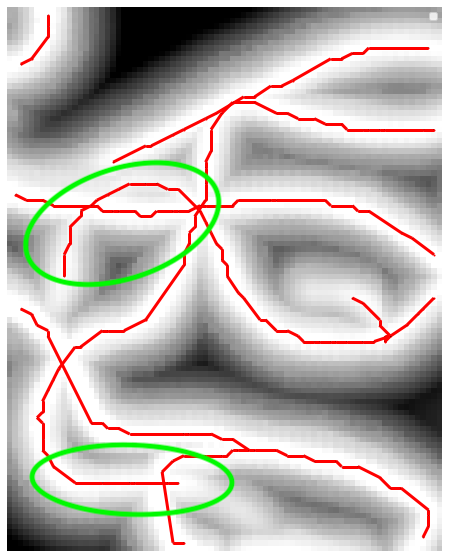} &
\includegraphics[height=0.1625\textwidth,angle=90]{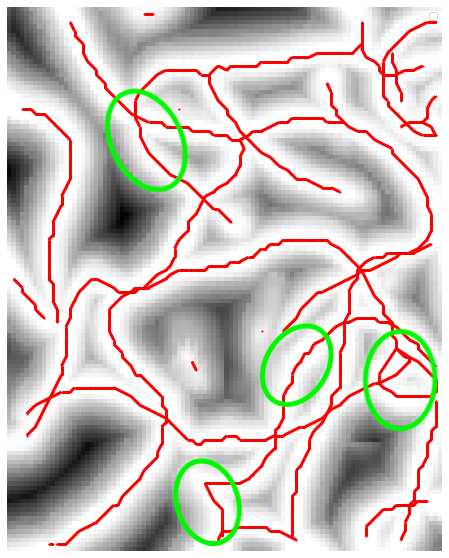} \\

\end{tabular}
\caption{ \label{fig:visual_combined}
Test predictions of different methods on three data sets. The green ellipses denote areas where training with the original annotations results in unwarranted breaks in the delineations whereas our approach does not.
}
\end{figure*}

% !TEX root = ../top.tex
% !TEX spellcheck = en-US

\begin{figure*}[!htb]
\centering
\begin{tabular}{@{}>{\centering\arraybackslash}m{0.02\textwidth} 
                @{}>{\centering\arraybackslash}m{0.1625\textwidth} 
                @{}>{\centering\arraybackslash}m{0.1625\textwidth}
                @{}>{\centering\arraybackslash}m{0.1625\textwidth} 
                @{}>{\centering\arraybackslash}m{0.1625\textwidth} 
                @{}>{\centering\arraybackslash}m{0.1625\textwidth}
                @{}>{\centering\arraybackslash}m{0.1625\textwidth} @{}}
& \multicolumn{2}{c}{\neurons{}} & \multicolumn{2}{c}{\neuronsmisal{}} & \multicolumn{2}{c}{\mra{}} \\

\rotatebox[origin=l]{90}{\footnotesize \emph{input + annot.}} &
\includegraphics[height=0.1625\textwidth,angle=90]{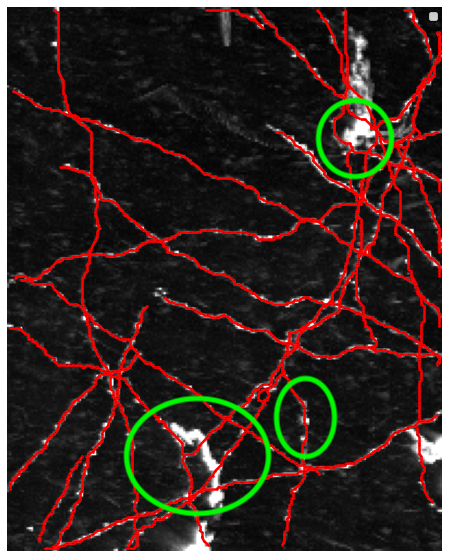} &
\includegraphics[height=0.1625\textwidth,angle=90]{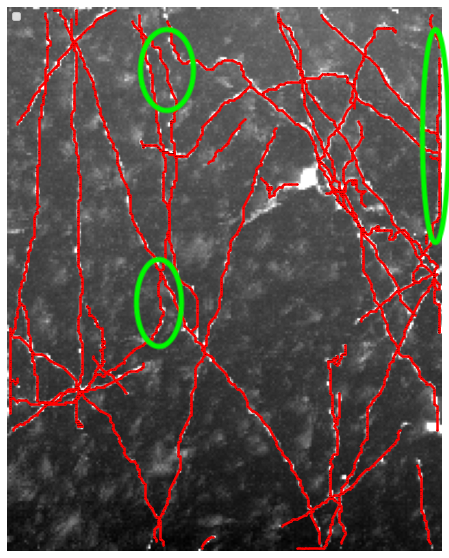} &
\includegraphics[height=0.1625\textwidth,angle=90]{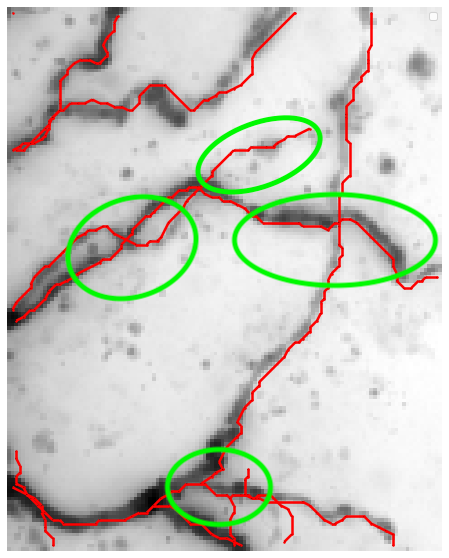} &
\includegraphics[height=0.1625\textwidth,angle=90]{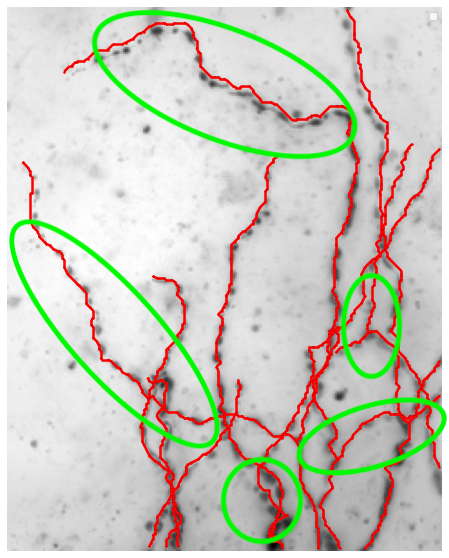} &
\includegraphics[height=0.1625\textwidth,angle=90]{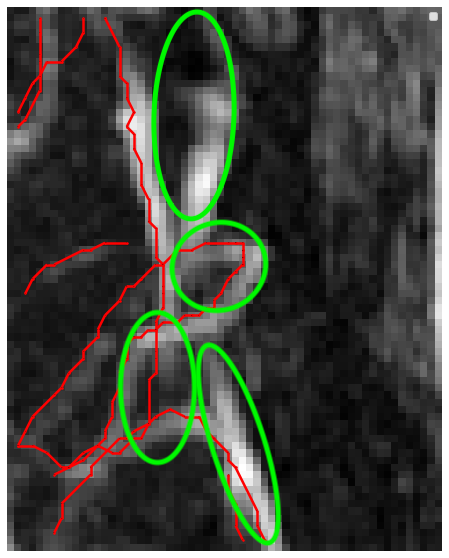} &
\includegraphics[height=0.1625\textwidth,angle=90]{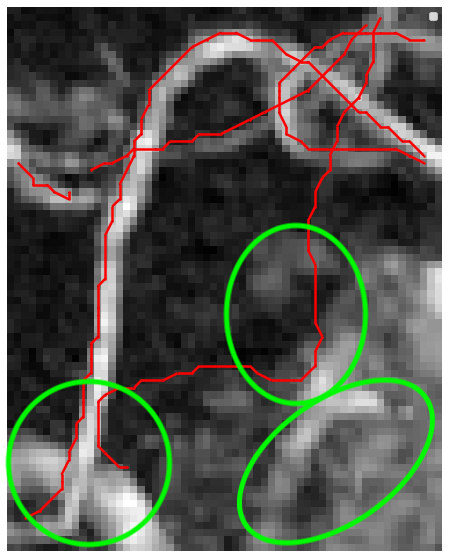} \\

{\rotatebox[origin=l]{90}{\footnotesize \COPLE{} }} &
\includegraphics[height=0.1625\textwidth,angle=90]{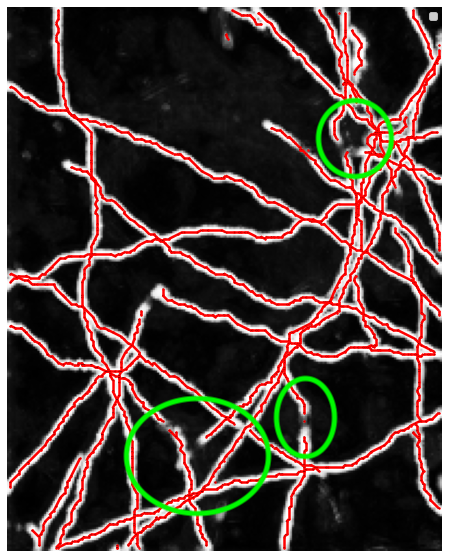} &
\includegraphics[height=0.1625\textwidth,angle=90]{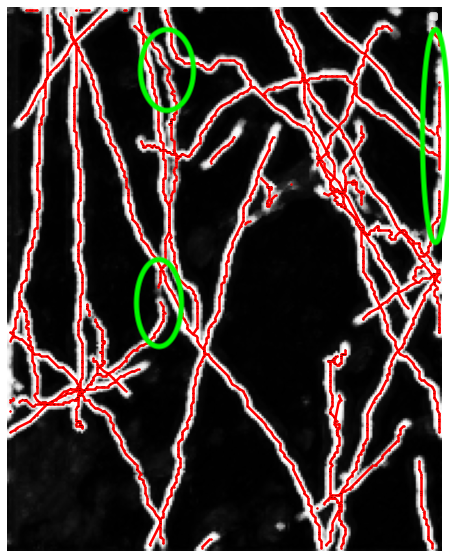} &
\includegraphics[height=0.1625\textwidth,angle=90]{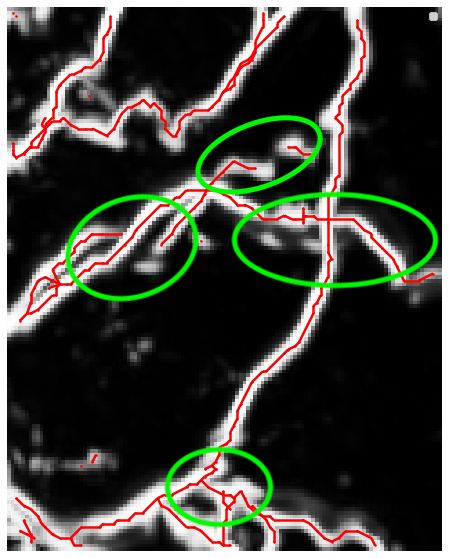} &
\includegraphics[height=0.1625\textwidth,angle=90]{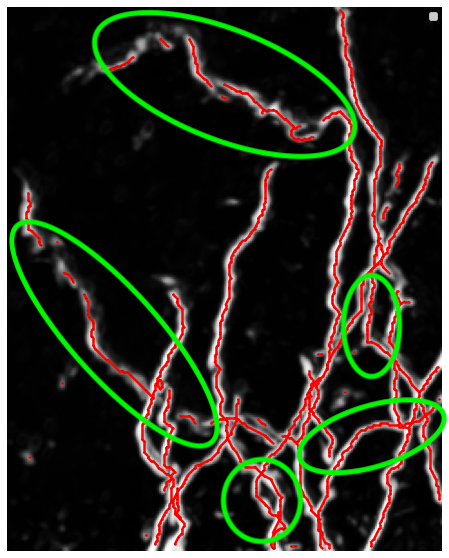} &
\includegraphics[height=0.1625\textwidth,angle=90]{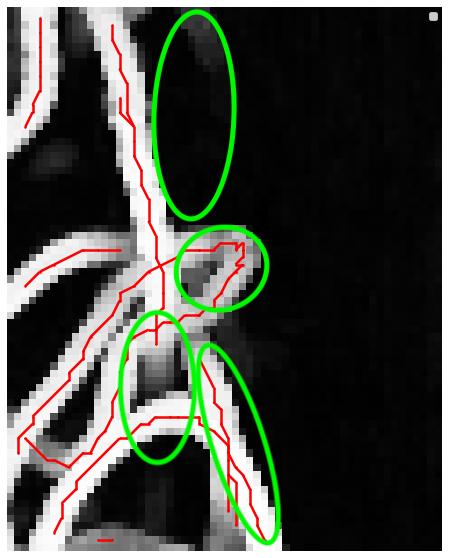} &
\includegraphics[height=0.1625\textwidth,angle=90]{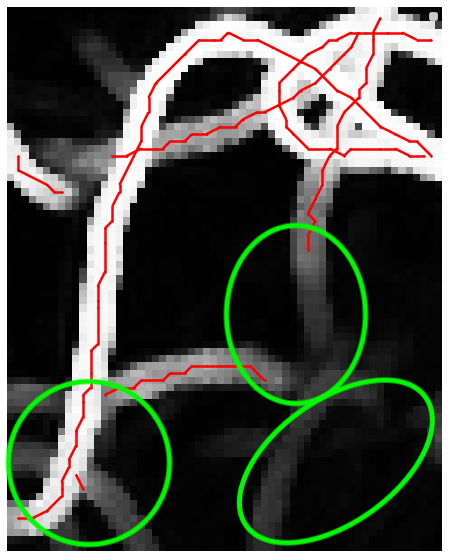} \\

\rotatebox[origin=l]{90}{\footnotesize \QAM{} } &
\includegraphics[height=0.1625\textwidth,angle=90]{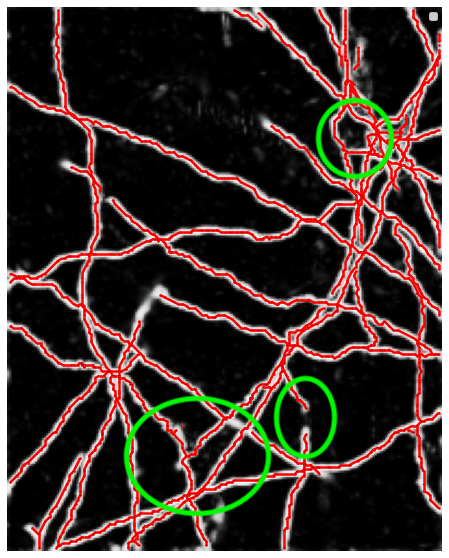} &
\includegraphics[height=0.1625\textwidth,angle=90]{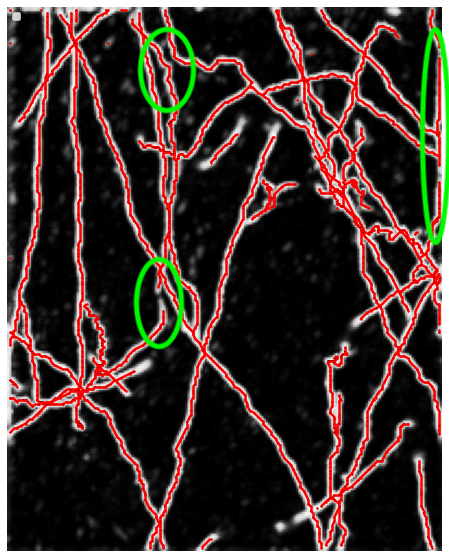} &
\includegraphics[height=0.1625\textwidth,angle=90]{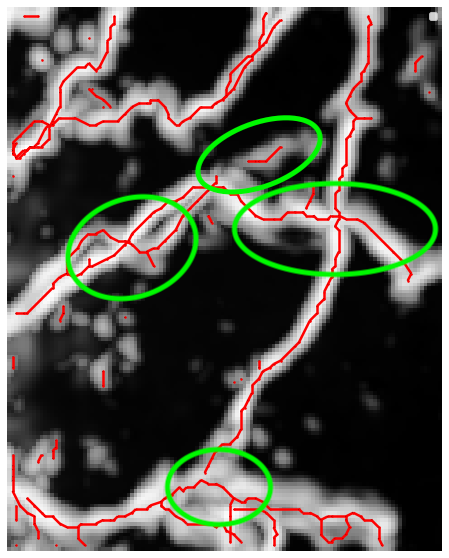} &
\includegraphics[height=0.1625\textwidth,angle=90]{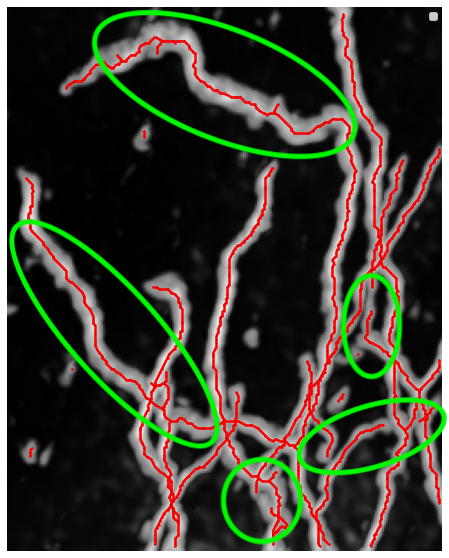} &
\includegraphics[height=0.1625\textwidth,angle=90]{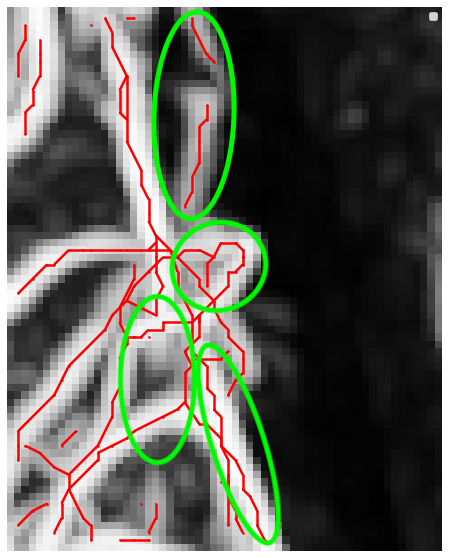} &
\includegraphics[height=0.1625\textwidth,angle=90]{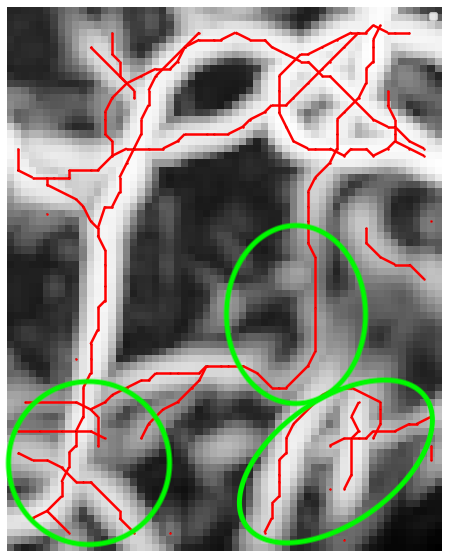} \\

\rotatebox[origin=l]{90}{\footnotesize \DS{} } &
\includegraphics[height=0.1625\textwidth,angle=90]{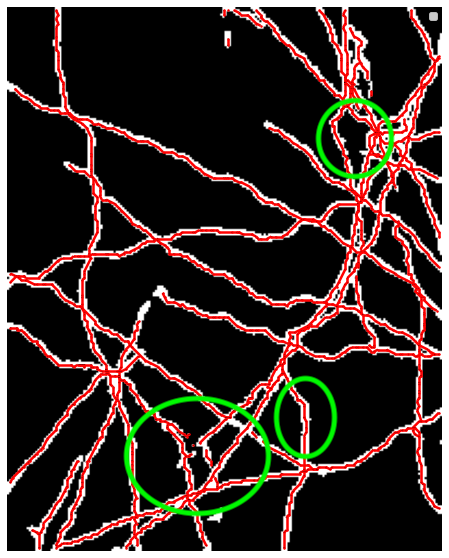} &
\includegraphics[height=0.1625\textwidth,angle=90]{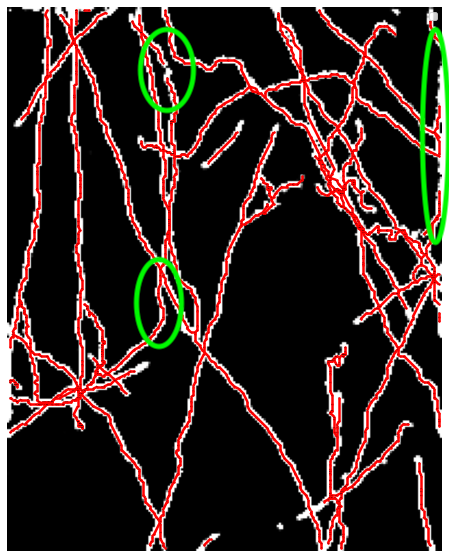} &
\includegraphics[height=0.1625\textwidth,angle=90]{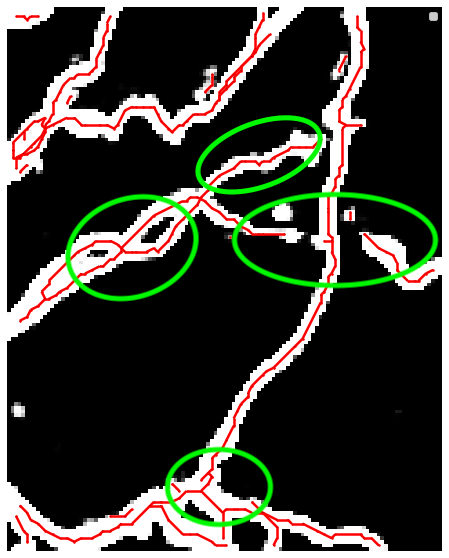} &
\includegraphics[height=0.1625\textwidth,angle=90]{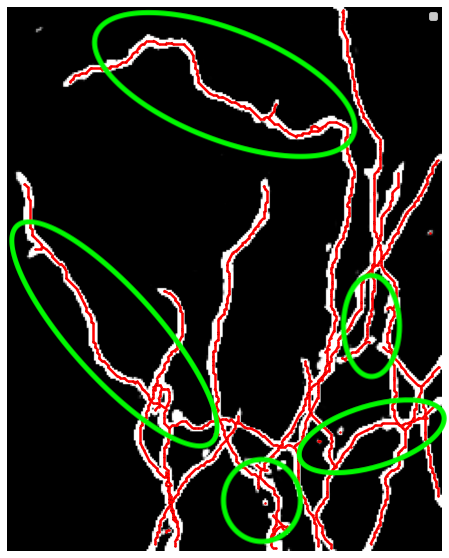} &
\includegraphics[height=0.1625\textwidth,angle=90]{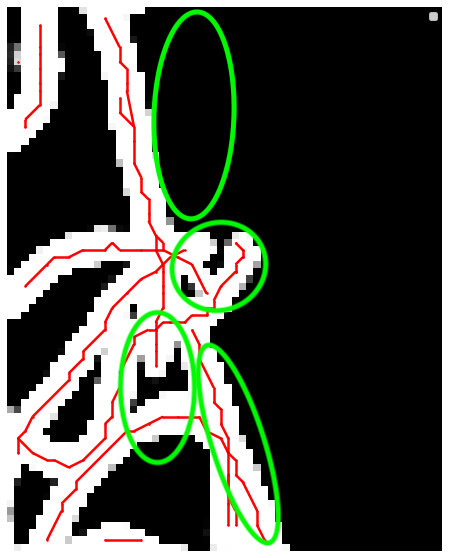} &
\includegraphics[height=0.1625\textwidth,angle=90]{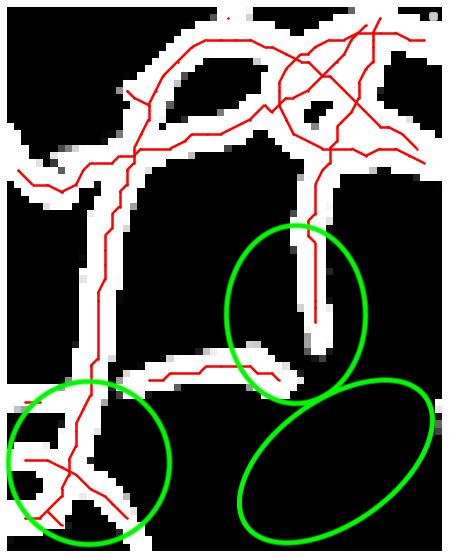} \\

\rotatebox[origin=l]{90}{\footnotesize \UNet{}+\snakeFast{} } &
\includegraphics[height=0.1625\textwidth,angle=90]{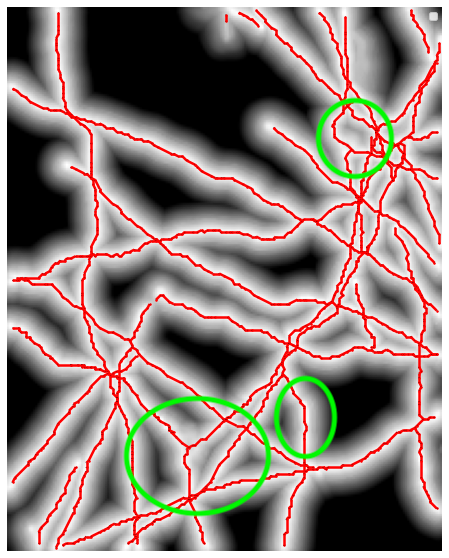} &
\includegraphics[height=0.1625\textwidth,angle=90]{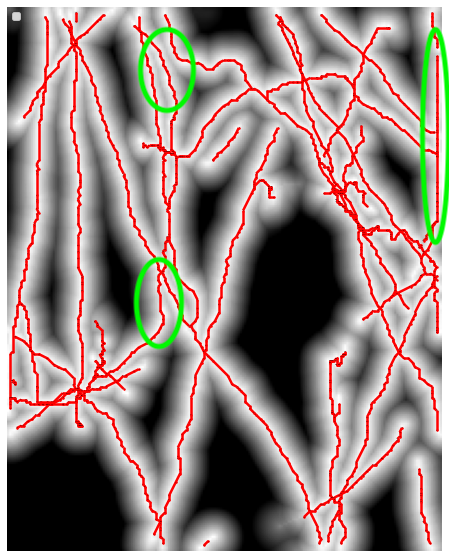} &
\includegraphics[height=0.1625\textwidth,angle=90]{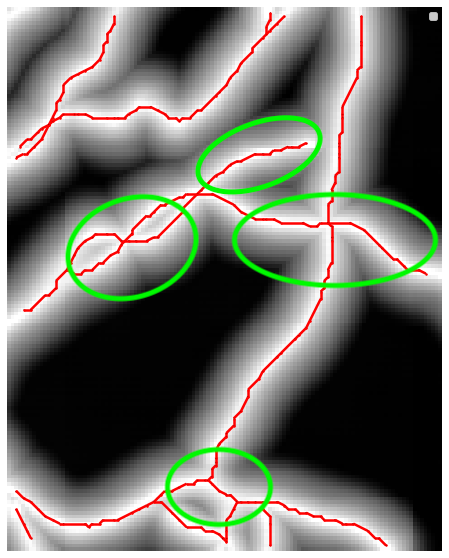} &
\includegraphics[height=0.1625\textwidth,angle=90]{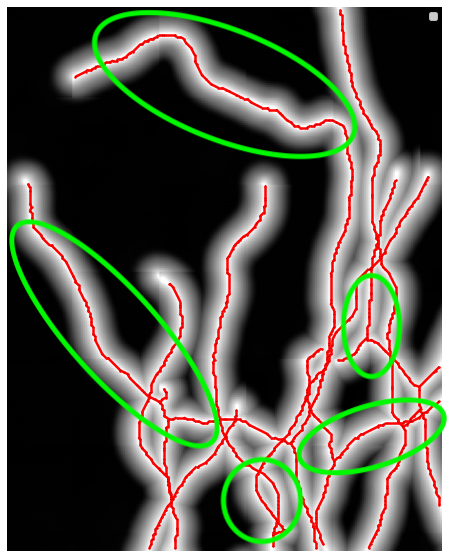} &
\includegraphics[height=0.1625\textwidth,angle=90]{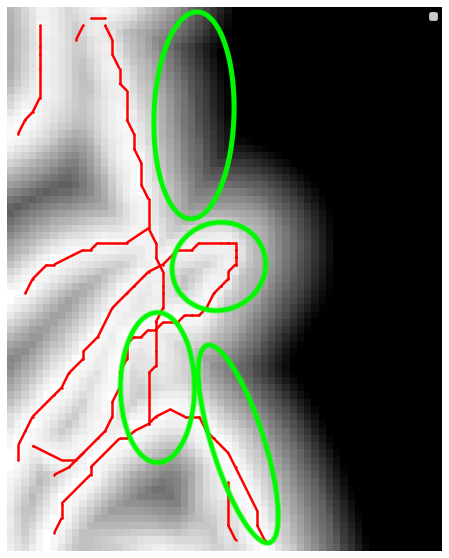} &
\includegraphics[height=0.1625\textwidth,angle=90]{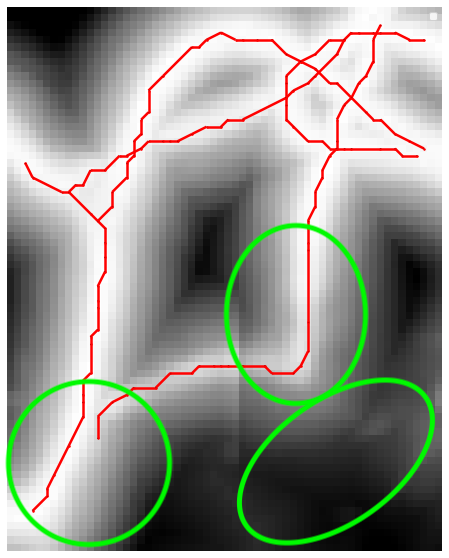} \\

\end{tabular}
\caption{ \label{fig:visual_sota}
Qualitative comparison of the results of \snakeFast{} to existing methods of training with noisy labels. The green ellipses denote areas where baselines result in unwarranted breaks in the delineations at test time whereas our approach does not.
}
\end{figure*}

Our contribution lies in the updating of the annotations and the loss function we use to achieve it, which should improve performance independently of any specific network architecture. To demonstrate this, we used two different architectures.
\begin{itemize}
\item \UNet{}. A 3D \UNet{}~\cite{Ronneberger15} with three max-pooling layers and two convolutional blocks. The first layer has 64 filters. Each convolution layer is followed by a batch-normalization and dropout with a probability of 0.15. During training, we randomly crop sub-volumes of size $96\times96\times96$ and flip them along each dimension with probability $0.5$. We combine them into batches of 8. 

\item \DRU{}. A recurrent architecture iteratively refining segmentation output 3 times~\cite{Wang19c}. The first layer has 64 filters. Each convolution layer is followed by a group-normalization and dropout with a probability of 0.15. During training, we randomly crop sub-volumes of size $96\times96\times96$ and flip them along each dimension with probability $0.5$. We combine them into batches of 4. To compute the loss function, we average the outputs of all 3 refinement steps. During testing, the output of the final step is used to evaluate performance. 
\end{itemize}
We trained both architectures in four different ways: by minimizing the Mean Squared Error to the original annotations, which we will refer to as \baseline{}, and by using the \snakeSimple{},  \snakeFull{},  and \snakeFast{} variants of our approach, described in Sections~\ref{sec:grad} and depicted by Fig.~\ref{fig:grads}. In all cases, we used Adam~\cite{Kingma15} with the learning rate set to $1e-4$, and a weight decay of $1e-4$. At test time, the predicted distance map were thresholded at 2 and skeletonized to obtain centerlines. To compute the \TLTS{} and \APLS{} scores, we converted them into graphs.

\subsection{Label Correction Baselines}
\label{sec:labelCorrect}

{
As noted in section~\ref{sec:related}, we do not know of other methods that deform the annotation graph during training, while maintaining its topology. However, there are methods designed to train deep nets using noisy annotations, where the noise is understood as flipping some pixel labels. In the following section, we compare our algorithm to three such methods:
}
\begin{itemize}
	\item {\COPLE{}.   A \UNet{} trained with the Noise Robust Dice Loss proposed in~\cite{Wang20i}.}
	\item {\QAM{}. An architecture with an auxiliary deep network to recognize annotations that might be wrong and downplay their importance during training~\cite{Zhu19}.}
	\item {\DS{}. A Siamese architecture and a training routine dedicated to enforcing equivariance of the network to deformations~\cite{Chatterjee20}.} 
\end{itemize}

\subsection{Comparative Evaluation.} 
\label{subs:results}

We present example reconstructions in \mk{Fig.~\ref{fig:visual_combined}} and Fig.~\ref{fig:visual_sota}. As shown in Tab.~\ref{tab:results_combined}, \snakeFull{} and \snakeFast{} outperform \baseline{} in \CCQ{} terms by a small margin, and in \APLS{} and \TLTS{} terms by a significantly larger one, which confirms that the main benefit of our loss is the improved connectivity of the predictions. As can be seen in Fig~\ref{fig:visual_combined}, our approach to training yields delineations with  fewer unwarranted breaks and longer uninterrupted curvilinear segments. 

On average \UNet{} and \DRU{} perform best when trained with \snakeFull{} and \snakeFast{}. However, \snakeFast{} requires three times less time per training iteration. \snakeSimple{} delivers a further 20-30\% speedup but incurs a clear performance drop. Crucially, these conclusions apply to both the \UNet{} and \DRU{} architectures. In fact, the performance gain resulting from switching from \baseline{} to \snakeFast{} is larger than the one resulting from changing from the simpler \UNet{} to the more sophisticated \DRU{} while retaining the standard \baseline{} approach to training.

In short, \snakeFast{} represents an excellent compromise between training time and performance. This being said, at test time, the run-time is the same no matter how the network was trained, because there is no alignment of annotations anymore. Hence, given sufficient computational resources, \snakeFull{} is also a valid option.

%We attribute this to the weakness of \snakeSimple{} described in section~\ref{sec:speedup}. The adjusted annotations optimize a snake objective function $S$, that can be computed efficiently. However, $S$ may fail to approximate the data objective $L$ sufficiently well, in which case the annotations adjusted by \snakeSimple{} fit the prediction less well than ones adjusted by \snakeFull{}, disrupting training. Even though the same snake objective $S$ is used in \snakeFast{}, this weakness is addressed by also backpropagating by the snake update procedure, which makes the prediction guide the snake to a pose in which $L$ is optimized. Even though, as shown in section~\ref{sec:speedup}, this may introduce artifacts in the prediction, for real data we did not observe any perturbations that negatively affected performance.

{
The bottom third of each part of Tab.~\ref{tab:results_combined} measures the performance of the methods designed to accommodate label noise, as described in Section~\ref{sec:labelCorrect}. Because they don't explicitly preserve annotation topology and we do, \UNet{} trained with \snakeFast{} outperform these methods in terms of the topology-aware scores but not necessarily in terms of the pixel-aware ones, which are note our main concern. 
} 

\subsection{Perfectly Accurate Annotations}
\label{sec:precise}

% !TEX root = ../top.tex
% !TEX spellcheck = en-US

\begin{table}
\caption{
Performance of a \UNet{} trained with the \baseline{} and with \snakeFast{} on the \synth{} data set with precise annotations.
%See section~\ref{subs:results} for description.
 \label{tab:results_synth}
}
\setlength{\tabcolsep}{3pt}
\center
\begin{tabular}{@{}l l  c c c  c c c  c  c c @{}}
 & & \multicolumn{3}{c}{Pixel-wise} & &  \multicolumn{2}{c}{Topology-aware} & & iter.\ t.\\
\cmidrule{3-5}
\cmidrule{7-8}
\cmidrule{10-10}
Arch. &  Method & Corr. & Compl.  & Qual.&   &APLS & TLTS & & s\\
\cmidrule{1-10}
\multirow{2}{*}{\UNet{}}
&\baseline{}   & 86.9 & 86.5 &{\bf 77.2}&&     92.8 &     89.0 && 2.8 \\ 
&\snakeFast{}  & 86.6 & 86.7 &     77.1 &&{\bf 93.2}&{\bf 89.3}&& 4.8 \\
\cmidrule{1-10}
\end{tabular}
\end{table}

% !TEX root = ../top.tex
% !TEX spellcheck = en-US

\begin{figure}[!htb]
\centering
\setlength{\tabcolsep}{0pt}
{
\begin{tabular}{ccc}
\includegraphics[width=0.33\columnwidth]{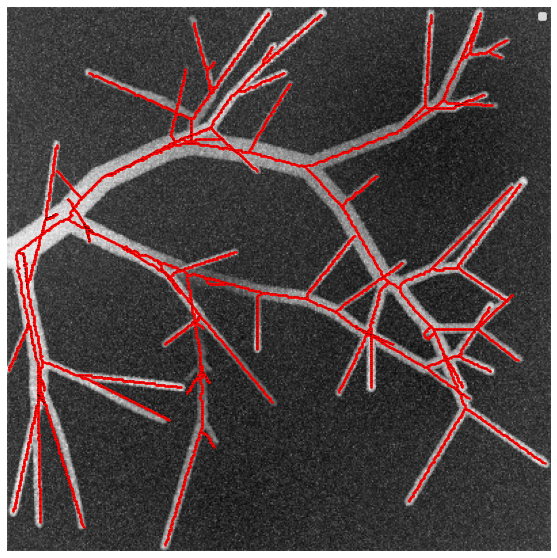} &
\includegraphics[width=0.33\columnwidth]{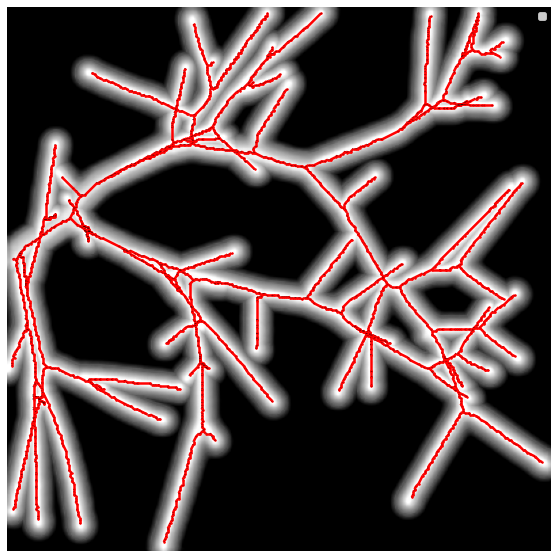} &
\includegraphics[width=0.33\columnwidth]{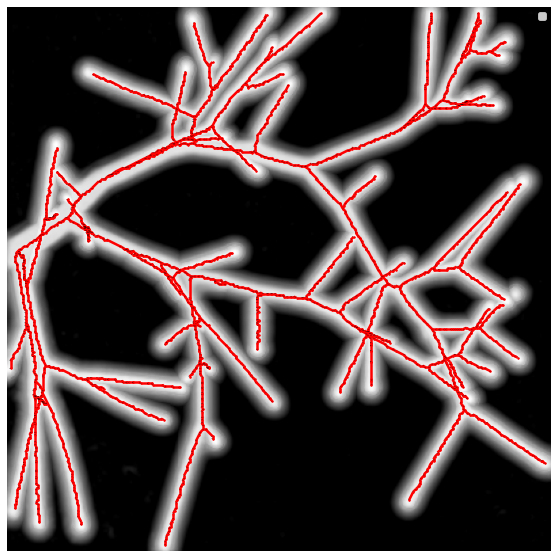}  \\
\scriptsize input and annotation & \scriptsize result of \baseline{}  & \scriptsize result of \snakeFast{} 
\end{tabular}
}
\caption{ \label{fig:visual_synth}
{
Results of training with the precise annotations of the \synth{} data set.
When the annotations are precise, \snakeFast{} performs as well as training with the \baseline{}.
}
}
\end{figure}

Having demonstrated that our loss function improves delineation results when the annotations lack spatial precision in Sec.~\ref{subs:results}, we now investigate its behavior when the annotation is precise. Since it is virtually impossible to precisely annotate 3D microscopy scans, we resort to synthetic data set \synth{}, which we generated using the VascuSynth algorithm~\cite{Hamarneh10,Jassi11} and its implementation~\cite{Zhang19}. The images are generated from vascular graphs, which we use as perfectly accurate annotations. We used twenty stacks for training and ten for testing, each of size $400 \times 400 \times 400$. Fig.~\ref{fig:visual_synth} shows the maximum-intensity projection of a test stack. The results, presented in Tab.~\ref{tab:results_synth}, confirm that, for perfectly accurate annotations, our method reduces to standard training with the MSE without incurring any performance drop.

\subsection{Increasing Annotation Inaccuracy}
\label{sec:varying}

% !TEX root = ../top.tex
% !TEX spellcheck = en-US

\begin{figure}[!htb]
	\setlength{\tabcolsep}{1pt}
	\centering
	\begin{tabular}{ccc} 
		\includegraphics[width=0.32\columnwidth]{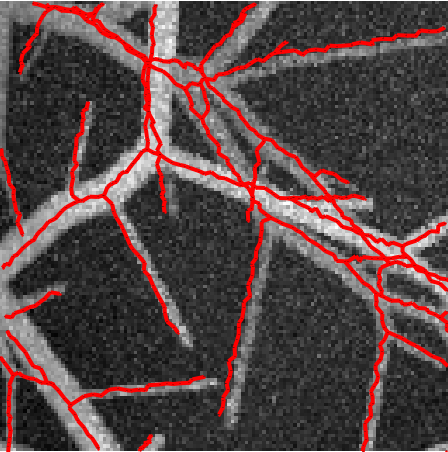} &
		\includegraphics[width=0.32\columnwidth]{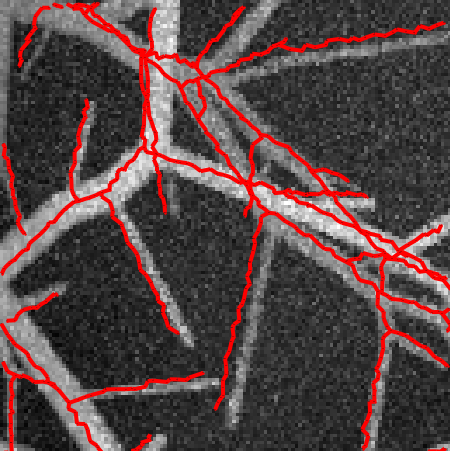} &
		\includegraphics[width=0.32\columnwidth]{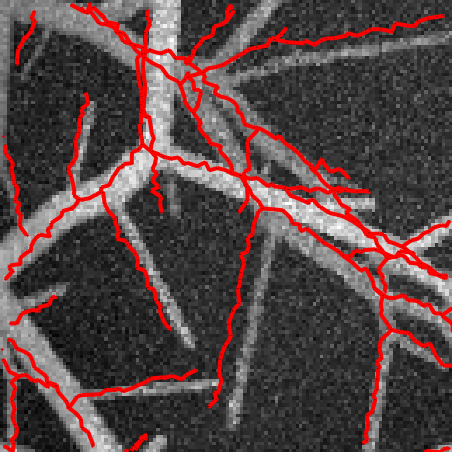}\\
		\scriptsize \emph{Level 1} & \scriptsize \emph{Level 2} & \scriptsize \emph{Level 3}
	\end{tabular}
	\vspace{-2mm}
	\caption{
\label{fig:teaser_noisy}
\change{
{\bf Annotation Deformation Levels.}  The deformation magnitude increases from left to right.
}
}
\end{figure}

% !TEX root = ../top.tex
% !TEX spellcheck = en-US

\begin{figure}[ht!]
	\setlength{\tabcolsep}{1pt}
	\centering
	%\begin{tabular}{ccc} 
	\begin{tabular}{cc} 
		\includegraphics[width=0.50\columnwidth]{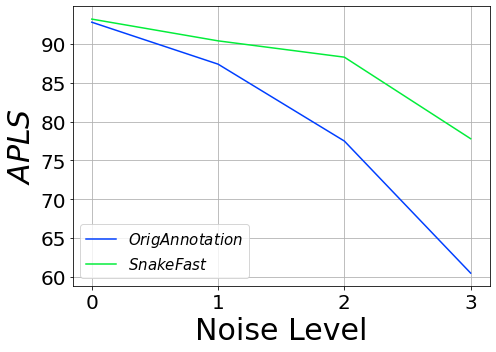} &
		\includegraphics[width=0.50\columnwidth]{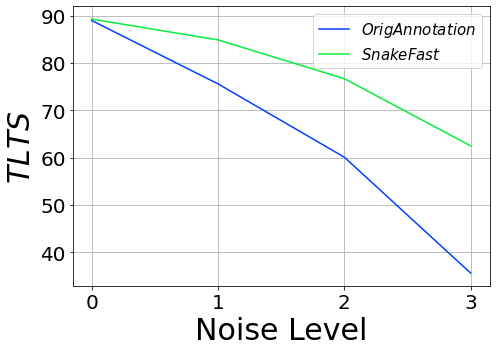} %&
		%\includegraphics[width=0.33\columnwidth]{fig/qual_slack2}
%                \\
%		\scriptsize (a)&\scriptsize (b) \\ %&(c)
	\end{tabular}
%	\vspace{-2mm}
\caption{
\label{fig:noise_effect}
\change{
{\bf Increasing the amount of deformation.} \APLS{}  and \TLTS{}  scores as a function of the deformation level. The \baseline{} scores decrease fast whereas those of  \snakeFast{} decrease much more slowly.}
}
\end{figure}

\change{
To investigate how increasing the level of inaccuracy of the annotations affects the performance of a \UNet{} trained with \snakeFast{}, we perturbed the annotations of the \synth{} data set.  We applied a random deformation field that varies  slowly across space to each annotation graph. We modulated its amplitude to change the level of inaccuracy. This produced three sets of annotations, as depicted by Fig.~\ref{fig:teaser_noisy}. We trained the network on each of them and present the results in Fig.~\ref{fig:noise_effect}. When the network is trained with \snakeFast{}, its connectivity-related scores degrade much slower than when trained using \baseline{}.
}

\subsection{Reducing Annotation Effort}
\label{sec:annotation_effort}

% !TEX root = ../top.tex
% !TEX spellcheck = en-US
\begin{table}
\caption{
Performance of \UNet{} trained using \snakeFast{} and \baseline{} on the \neurons{} data set with very coarse annotations.
Performance of \UNet{} trained using the precise annotations shown for reference.
%See section~\ref{subs:ablation} for description.
 \label{tab:coarse}
}
\setlength{\tabcolsep}{3pt}
\center

\begin{tabular}{@{}l c c  c c  c c c  c  @{}}
&  & \multicolumn{3}{c}{Pixel-wise} & &  \multicolumn{2}{c}{Topology-aware} \\
\cmidrule{3-5}
\cmidrule{7-8}
Annot. & Method & Corr. & Compl.  & Qual.&   &APLS & TLTS \\
\cmidrule{1-8}
\multirow{2}{*}{coarse} 
& \baseline{}  & 85.2 & 67.6 &     60.4 &&     46.5 &     50.9  \\ 
& \snakeFast{} & 97.6 & 87.0 &{\bf 85.3}&&{\bf 66.8}&{\bf 73.5} \\
%\cmidrule{1-8}
\rule{0pt}{3ex}\multirow{2}{*}{precise}
& \baseline{}  & 98.9 & 91.3 &     90.4 &&     80.3 &     80.9 \\ 
& \snakeFast{} & 98.7 & 95.0 &{\bf 93.8}&&{\bf 91.1}&{\bf 85.9} \\
        \cmidrule{1-8}
\end{tabular}

%\begin{tabular}{@{}l l c c c  c c c  c  c c @{}}
%\cmidrule{3-10}
%&  & \multicolumn{3}{c}{Pixel-wise} & &  \multicolumn{2}{c}{Topology-aware} & & iter.\ t.\\
%\cmidrule{3-5}
%\cmidrule{7-8}
%\cmidrule{10-10}
%shift  & & Corr. & Compl.  & Qual.&   &APLS & TLTS & & s\\
%\cmidrule{1-10}
%\multirow{2}{*}{0} 
%& \baseline{}  &     ??.? &     ??.? &     ??.? &&     ??.? &     ??.? && ?.? \\ 
%& \snakeFast{} &     ??.? &     ??.? &     ??.? &&     ??.? &     ??.? && ?.? \\[2mm] 
%\multirow{2}{*}{1} 
%& \baseline{}  &     ??.? &     ??.? &     ??.? &&     ??.? &     ??.? && ?.? \\ 
%& \snakeFast{} &     ??.? &     ??.? &     ??.? &&     ??.? &     ??.? && ?.? \\[2mm] 
%\multirow{2}{*}{3} 
%& \baseline{}  &     ??.? &     ??.? &     ??.? &&     ??.? &     ??.? && ?.? \\ 
%& \snakeFast{} &     ??.? &     ??.? &     ??.? &&     ??.? &     ??.? && ?.? \\[2mm] 
%\multirow{2}{*}{7} 
%& \baseline{}  &     ??.? &     ??.? &     ??.? &&     ??.? &     ??.? && ?.? \\ 
%& \snakeFast{} &     ??.? &     ??.? &     ??.? &&     ??.? &     ??.? && ?.? \\[2mm] 
%\cmidrule{1-10}
%\end{tabular}
\end{table}

% !TEX root = ../top.tex
% !TEX spellcheck = en-US

\begin{figure}[ht]
\setlength{\tabcolsep}{1pt}
\centering
\begin{tabular}{ccc} 
\includegraphics[width=0.33\columnwidth]{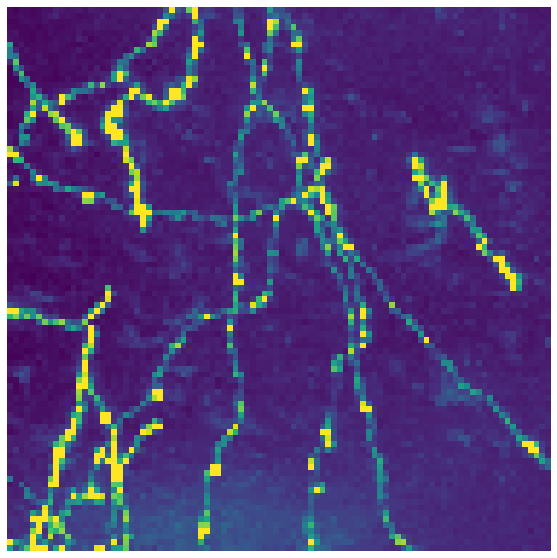} &
\includegraphics[width=0.33\columnwidth]{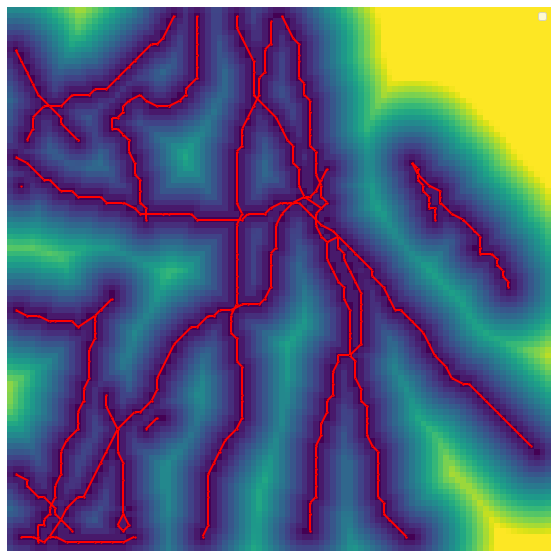} &
\includegraphics[width=0.33\columnwidth]{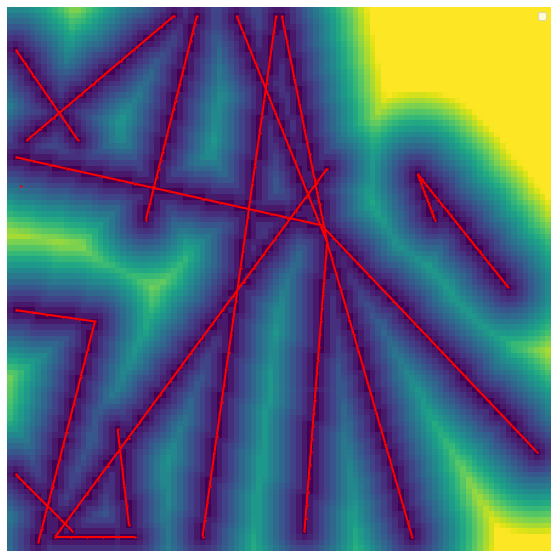}\\
\scriptsize (a)& \scriptsize (b)& \scriptsize (c)
\end{tabular}
\vspace{-2mm}
\caption{{\bf Coarse Annotations}
\label{fig:teaser_easy}
(a) Training image of a neurite (b) Distance map obtained from original annotation overlaid in red (c) Distance map obtained from coarse annotation overlaid in red. Coarse annotations are obtained by connecting neurite end points and bifurcations with straight lines, and are easier to perform than full annotations. 
}
\end{figure}

% !TEX root = ../top.tex
% !TEX spellcheck = en-US

\begin{figure}[ht]
\centering
\setlength{\tabcolsep}{0pt}
\begin{tabular}{@{} ccc @{}}

\includegraphics[height=0.33\columnwidth,angle=90]{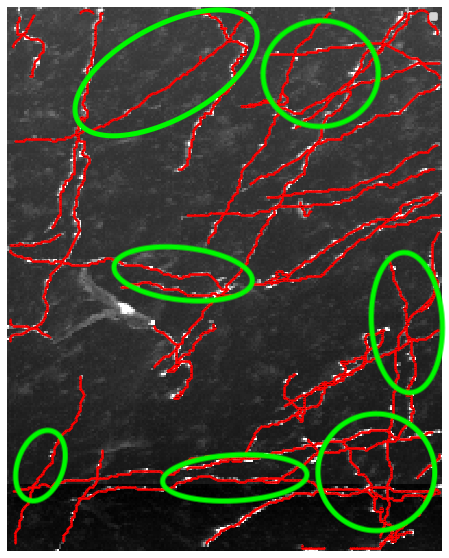} &
\includegraphics[height=0.33\columnwidth,angle=90]{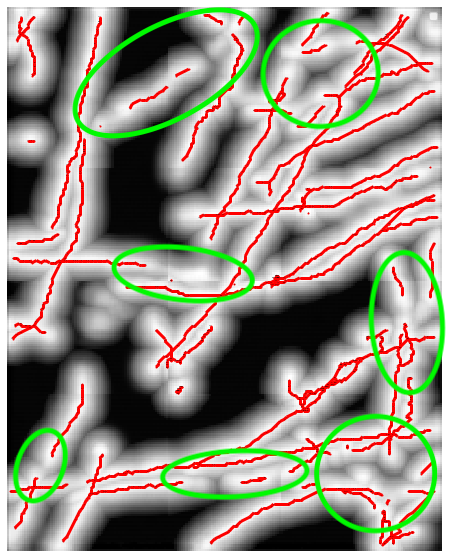} &
\includegraphics[height=0.33\columnwidth,angle=90]{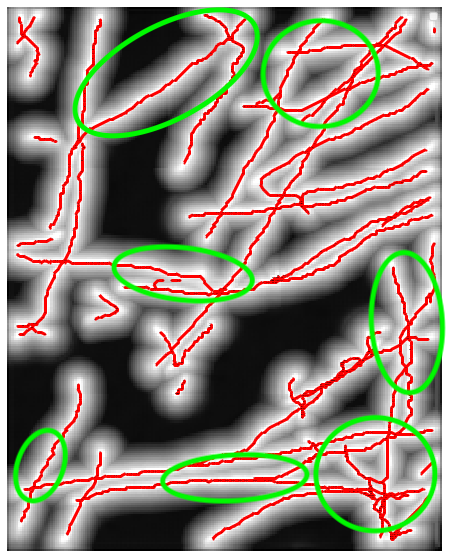}  \\

\includegraphics[height=0.33\columnwidth,angle=90]{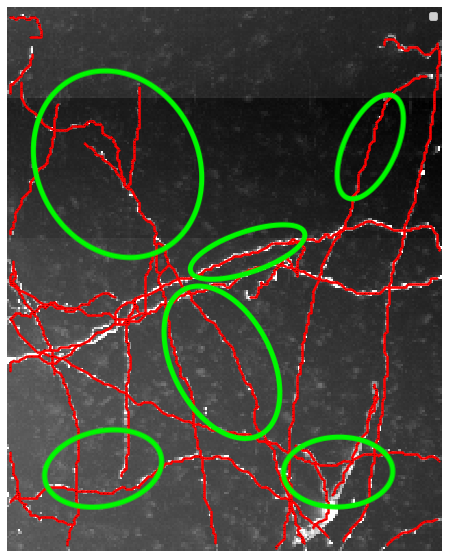} &
\includegraphics[height=0.33\columnwidth,angle=90]{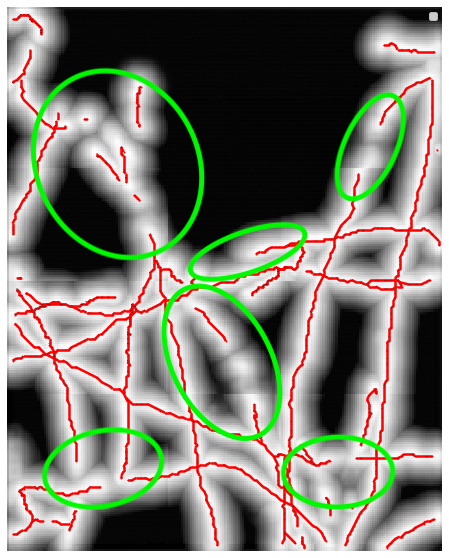} &
\includegraphics[height=0.33\columnwidth,angle=90]{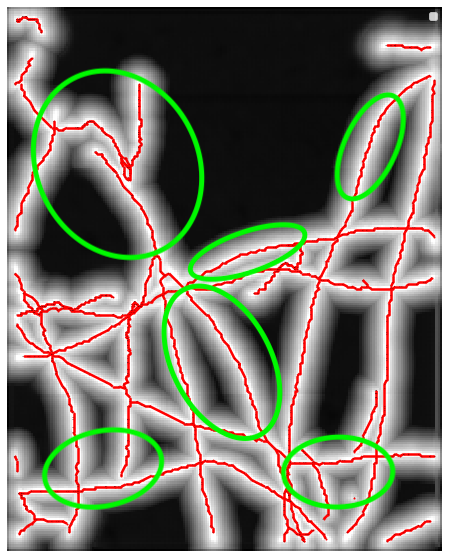}  \\

\scriptsize input & \scriptsize \baseline{} & \scriptsize \snakeFast{} 

\end{tabular}
\caption{ \label{fig:visual_easy}
Results of training a \UNet{} with \baseline{} and \snakeFast{} on the \neurons{} data set with easy annotations. 
}
\end{figure}

The robustness of \snakeFast{} to deviations in the annotation inspired us to ask another question: Can this loss function be used to train deep networks with annotations that are simplified to the point where they become much easier, faster, and therefore cheaper to obtain?
To answer this,
we trained the \UNet{} with \snakeFast{} and \baseline{} on the \neurons{} data set with very coarse annotations.
We obtained them by connecting neurite branching- and end-points with straight lines, as shown in Fig.~\ref{fig:teaser_easy}. 
The results are presented in Tab.~\ref{tab:coarse}.
As expected, training on the coarse annotations without adjusting them results in a significant performance drop as compared to training on precise annotations.
Switching from precise to coarse annotations still incurs a performance drop when using \snakeFast{}, but a much smaller one than when using the baseline. Visual inspection of the resulting segmentations, shown in Fg.~\ref{fig:visual_easy}, leads us to conclude that, for tasks where a compromise between accuracy and annotation cost is acceptable, using the easy annotations together with \snakeFast{} is a viable alternative to the classical approach.

%The gap between performance of \snakeFast{} on the coarse and precise annotations shows the main limitation of our algorithm.
%The snake converges to a local optimum, and therefore it may fail to correct very large annotation displacements.

\subsection{Ablation Studies.}
\label{subs:ablation}

To investigate the impact of hyper-parameters of our method on performance, we run the following ablation studies.
% !TEX root = ../top.tex
% !TEX spellcheck = en-US
\begin{table}
\caption{
Performance of \UNet{} trained using \snakeFast{} on the \neurons{} data set when varying the elasticity and spring term coefficients. 
%See section~\ref{subs:ablation} for description.
 \label{tab:alphabeta}
}
\setlength{\tabcolsep}{3pt}
\center
\begin{tabular}{@{}l l  c c c  c c c  c  c c @{}}
%\cmidrule{3-10}
 & & \multicolumn{3}{c}{Pixel-wise} & &  \multicolumn{2}{c}{Topology-aware} & & iter.\ t.\\
\cmidrule{3-5}
\cmidrule{7-8}
\cmidrule{10-10}
& & Corr. & Compl.  & Qual.&   &APLS & TLTS & & s\\
\cmidrule{1-10}
\multirow{4}{*}{$\alpha=$1e-2}
& $\beta=$1e-4 &     99.0 &     94.5 &     93.5 &&     88.1 &     84.8 && 5.2 \\ 
& $\beta=$1e-3 &     98.7 &     95.0 &{\bf 93.8}&&{\bf 91.1}&{\bf 85.9}&& 5.2 \\
& $\beta=$1e-2 &     98.4 &     94.0 &     92.7 &&     85.1 &     84.3 && 5.2 \\
& $\beta=$1e-1 &     98.9 &     93.8 &     92.8 &&     83.8 &     84.1 && 5.2 \\
\cmidrule{1-10}
$\alpha=$1e-4 &
\multirow{4}{*}{$\beta=$1e-3}
                &    98.4 &     92.9 &     91.5 &&     86.6 &     83.0 && 5.2 \\ 
$\alpha=$1e-3 & &    99.0 &     94.2 &     93.4 &&     85.3 &     84.4 && 5.2 \\
$\alpha=$1e-2 & &    98.7 &     95.0 &{\bf 93.8}&&{\bf 91.1}&{\bf 85.9}&& 5.2 \\
$\alpha=$1e-1 & &    98.7 &     94.3 &     93.1 &&     79.8 &     82.5 && 5.2 \\
\cmidrule{1-10}
\end{tabular}
\end{table}

\subsubsection{Regularization terms}
The regularization term $R$ of Eq.~\ref{eq:reg} is the sum of a spring term, weighted by a coefficient $\alpha$, and an elasticity term, weighted by a coefficient $\beta$.
To investigate their influence on performance, we varied $\alpha$ and $\beta$ and trained our \UNet{} on the \neurons{} data set.
The results are presented in Tab.~\ref{tab:alphabeta}. The best results are attained with relatively low values of both terms. Higher values of the spring term, originally proposed for closed contours, effectively reguralize loopy topologies, but when used on tree-shaped structures, representing blood vessels and neuronal processes, tend to shorten the reconstructed neurites and vessels. Higher values of the elasticity term make it more difficult to fit irregular trajectories of neurites, like the ones shown in Fig.~\ref{fig:visual_combined}.

% !TEX root = ../top.tex
% !TEX spellcheck = en-US
\begin{table}
\caption{
Performance of \UNet{} trained using \snakeFast{} on the \neurons{} data set when varying the inverse stepsize, together with the number of snake updates used in every training iteration and the resulting iteration time.
%See section~\ref{subs:ablation} for description.
 \label{tab:stepsize}
}
\setlength{\tabcolsep}{3pt}
\center
\begin{tabular}{@{}l c c c  c c c  c  c c c c@{}}
%\cmidrule{2-11}
 & \multicolumn{3}{c}{Pixel-wise} & &  \multicolumn{2}{c}{Topology-aware} & & no steps && iter.\ t. \\
\cmidrule{2-4}
\cmidrule{6-7}
\cmidrule{9-9}
\cmidrule{11-11}
& Corr. & Compl.  & Qual.&   &APLS & TLTS && && s\\
\cmidrule{1-11}
$\gamma=100$  &     98.8 &     94.5 &     93.4 &&     90.9 &     85.8 && 80 && 6.3 \\ 
$\gamma=10$   &     98.7 &     95.0 &{\bf 93.8}&&{\bf 91.1}&{\bf 85.9}&& 10 && 5.2 \\
%$\gamma=1$ &  38.5&     4.9 & 4.6 &&  1.4  &  3.1  && 10 && 5.2\\
$\gamma=1$    &  \multicolumn{6}{c}{ --- the snake diverged --- }     && 10 && 5.2 \\
\cmidrule{1-11}
\end{tabular}
\end{table}

\subsubsection{Step size for snake update}
As explained in section~\ref{sec:netsnakes}, the snake update iteration has a parameter $\gamma$, called viscosity, that acts as an inverse step size. 
We report the results of changing $\gamma$ in Tab.~\ref{tab:stepsize}.
Low viscosity results in large step size and can make the snake update procedure diverge, which we observed for $\gamma=1$.
On the other hand, high viscosity corresponds to small step size and increases the risk that the snake does not converge within the preset number of iterations.
With $\gamma=100$, we needed to increase the number of snake updates from 10 to 80 to ensure convergence.
This also increased the iteration time by one second.
$\gamma=10$ made the snake converge within 10 updates, while also resulting in marginally higher performance than $\gamma=100$. 

% !TEX root = ../top.tex
% !TEX spellcheck = en-US

\begin{table}
\caption{
Performance of deep nets trained with \MAE{} and \MSE{} costs on the \neurons{} data set and the time needed for single training iteration.
%See section~\ref{subs:results} for description.
 \label{tab:results_brain_mae}
}
\setlength{\tabcolsep}{3pt}
\center
\begin{tabular}{@{}l l  c c c  c c c  c  c c @{}}
%\cmidrule{3-10}
 & & \multicolumn{3}{c}{Pixel-wise} & &  \multicolumn{2}{c}{Topology-aware} & & iter.\ t.\\
\cmidrule{3-5}
\cmidrule{7-8}
\cmidrule{10-10}
Cost &  Method & Corr. & Compl.  & Qual.&   &APLS & TLTS & & s\\
\cmidrule{1-10}
\rule{0pt}{0ex}
\multirow{2}{*}{\MAE{}}
&\baseline{}   & 98.6 & 91.2 &     90.1 &&     81.4 &     80.5 && 2.8 \\ 
&\snakeFast{}  & 98.8 & 94.6 &{\bf 93.4}&&{\bf 89.9}&{\bf 85.8}&& 5.2 \\
%\cmidrule{1-10} 
\rule{0pt}{3ex}
\multirow{2}{*}{\MSE{}}
&\baseline{}   & 98.9 & 91.3 &     90.4 &&     80.3 &     80.9 && 2.8 \\ 
&\snakeFast{}  & 98.7 & 95.0 &{\bf 93.8}&&{\bf 91.1}&{\bf 85.9}&& 5.2 \\
\cmidrule{1-10}
\end{tabular}
\end{table}

\subsubsection{L1 vs L2 distance}
{
We also verified the performance of a \UNet{} trained with \snakeFast{} when changing the loss data term from Mean Squared Error to Mean Absolute Error. 
The results, shown in Tab.~\ref{tab:results_brain_mae} show very slight advantage of MSE, possibly due to a gradient profile that prioritizes penalizing higher errors.
}

%\input{fig/visual_brain}
%\input{fig/visual_misaligned}
%\input{fig/visual_mra}

%\input{fig/teaser_noise}
%\input{fig/visual_noise}
%\input{fig/slack_effect}

% !TEX root = ../top.tex
% !TEX spellcheck = en-US

\section{Conclusion and future work}

We have proposed a method that accounts for the inevitable inaccuracies in manual annotations of curvilinear 3D structures, such as neurites and blood vessels, in 3D image stacks. It leverages on the network snake formalism to define a loss function that simultaneously trains the deep network to produce the delineation and adjusts the initially imprecise annotations. 

Our approach does not depend on the specific network architecture we use. Hence, its effectiveness suggests that handling such imprecisions may be even more important than refining the network architecture, which is something that has been largely neglected in the literature.

In future work, we will investigate the extension our approach to segmenting surfaces, like cell membranes in electron microscopy scans. 

%use of even coarser annotations than the one we have used so far, so as to decrease the workload of the annotators as much as possible. Ideally, they should be able to provide only the nodes of the connectivity graph, that is, the places where there curvilinear structures meet or bifurcate, without having to trace the actual path between them. \PF{Sounds great but how are you going to achieve that? It could be done by using dynamic programming to trace between two nodes. But then do we really need your technique because the paths will be fairly accurate then?}

%we plan to use our loss function for training deep nets on very coarse annotations, that can be produced more quickly than the usual, more detailed ones. We will then investigate the potential of this technique to decrease the annotation effort needed to deploy a deep-learning-based system in a practical scenario. 

 \bibliographystyle{plain}
%\bibliography{cvlab-bib/biomed,cvlab-bib/vision,cvlab-bib/string} 
 \bibliography{string,vision,learning,biomed,misc}

\begin{thebibliography}{10}

\bibitem{Acuna19}
D.~Acuna, A.~Kar, and S.~Fidler.
\newblock {Devil is in the Edges: Learning Semantic Boundaries from Noisy
  Annotations}.
\newblock In {\em Conference on Computer Vision and Pattern Recognition}, 2019.

\bibitem{Bengio13a}
Y.~Bengio, A.~Courville, and P.~Vincent.
\newblock {Representation Learning: A Review and New Perspectives}.
\newblock {\em IEEE Transactions on Pattern Analysis and Machine Intelligence},
  2013.

\bibitem{Breitenreicher13}
D.~Breitenreicher, M.~Sofka, S.~Britzen, and S.K. Zhou.
\newblock {Hierarchical Discriminative Framework for Detecting Tubular
  Structures in {3D} Images}.
\newblock In {\em Conference on Medical Image Computing and Computer Assisted
  Intervention}, pages 328--340, 2013.

\bibitem{Bullitt05}
E.~Bullitt, D.~Zeng, G.~Gerig, S.~Aylward, S.~Joshi, J.~Smith, W.~Lin, and
  M.~Ewend.
\newblock {Vessel Tortuosity and Brain Tumor Malignancy: A Blinded Study}.
\newblock {\em Acad Radiol}, 12(10):1232--1240, October 2005.

\bibitem{Butenuth12}
M.~Butenuth and C.~Heipke.
\newblock {Network Snakes: Graph-Based Object Delineation with Active Contour
  Models}.
\newblock {\em Machine Vision and Applications}, 23(1):91--109, 2012.

\bibitem{Chatterjee20}
S.~Chatterjee, K.~Prabhu, M.~Pattadkal, G.~Bortsova, C.~Sarasaen, F.~Dubost,
  H.~Mattern, M.~de~Bruijne, O.~Speck, and A.~N{\"u}rnberger.
\newblock {Ds6, Deformation-Aware Semi-Supervised Learning: Application to
  Small Vessel Segmentation with Noisy Training Data}.
\newblock In {\em arXiv Preprint}, 2020.

\bibitem{Cheng19}
D.~Cheng, R.~Liao, S.~Fidler, and R.~Urtasun.
\newblock {DARNet: Deep Active Ray Network for Building Segmentation}.
\newblock In {\em Conference on Computer Vision and Pattern Recognition}, 2019.

\bibitem{Cootes01}
T.~F. Cootes, G.~J. Edwards, and C.~J. Taylor.
\newblock {Active Appearance Models}.
\newblock {\em IEEE Transactions on Pattern Analysis and Machine Intelligence},
  23(6), June 2001.

\bibitem{Doersch17}
C.~Doersch and A.~Zisserman.
\newblock {Multi-Task Self-Supervised Visual Learning}.
\newblock In {\em International Conference on Computer Vision}, October 2017.

\bibitem{Frangi98}
A.F. Frangi, W.J. Niessen, K.L. Vincken, and M.A. Viergever.
\newblock {Multiscale Vessel Enhancement Filtering}.
\newblock {\em Lecture Notes in Computer Science}, 1496:130--137, 1998.

\bibitem{Fua96f}
P.~Fua.
\newblock {Model-Based Optimization: Accurate and Consistent Site Modeling}.
\newblock In {\em International Society for Photogrammetry and Remote Sensing},
  July 1996.

\bibitem{Fua90}
P.~Fua and Y.~G. Leclerc.
\newblock {Model Driven Edge Detection}.
\newblock {\em Machine Vision and Applications}, 3:45--56, 1990.

\bibitem{Ganin14a}
Y.~Ganin and V.~Lempitsky.
\newblock {N4-Fields: Neural Network Nearest Neighbor Fields}.
\newblock In {\em Asian Conference on Computer Vision}, pages 536--551, 2014.

\bibitem{Hamarneh10}
G.~Hamarneh and P.~Jassi.
\newblock {Vascusynth: Simulating Vascular Trees for Generating Volumetric
  Image Data with Ground Truth Segmentation and Tree Analysis}.
\newblock {\em Computerized Medical Imaging and Graphics}, 34(8):605--616,
  2010.

\bibitem{Hatamizadeh20}
A.~Hatamizadeh, D.~Sengupta, and D.~Terzopoulos.
\newblock {End-To-End Trainable Deep Active Contour Models for Automated Image
  Segmentation: Delineating Buildings in Aerial Imagery}.
\newblock In {\em European Conference on Computer Vision}, 2020.

\bibitem{Huang09}
X.~Huang and L.~Zhang.
\newblock {Road Centreline Extraction from High-Resolution Imagery Based on
  Multiscale Structural Features and Support Vector Machines}.
\newblock {\em International Journal of Remote Sensing}, 30:1977--1987, 2009.

\bibitem{Jassi11}
P.~Jassi and G.~Hamarneh.
\newblock {Vascusynth: Vascular Tree Synthesis Software}.
\newblock {\em Insight Journal}, January-June:1--12, 2011.

\bibitem{Kass88}
M.~Kass, A.~Witkin, and D.~Terzopoulos.
\newblock {Snakes: Active Contour Models}.
\newblock {\em International Journal of Computer Vision}, 1(4):321--331, 1988.

\bibitem{Kingma15}
D.~P. Kingma and J.~Ba.
\newblock {Adam: {A} Method for Stochastic Optimisation}.
\newblock In {\em International Conference on Learning Representations}, 2015.

\bibitem{Kozinski20}
M.~Kozi{\'n}ski, A.~Mosinska, M.~Salzmann, and P.~Fua.
\newblock {Tracing in 2D to Reduce the Annotation Effort for 3D Deep
  Delineation of Linear Structures}.
\newblock {\em Medical Image Analysis}, 60, 2020.

\bibitem{Law08}
M.~Law and A.~Chung.
\newblock {Three Dimensional Curvilinear Structure Detection Using Optimally
  Oriented Flux}.
\newblock In {\em European Conference on Computer Vision}, pages 368--382,
  2008.

\bibitem{Maninis16}
K.K. Maninis, J.~Pont-Tuset, P.~Arbel\'{a}ez, and L.~Van Gool.
\newblock {Deep Retinal Image Understanding}.
\newblock In {\em Conference on Medical Image Computing and Computer Assisted
  Intervention}, pages 140--148, 2016.

\bibitem{Marcos18}
D.~Marcos, D.~Tuia, B.~Kellenbergerg, and R.~Urtasun.
\newblock {Learning Deep Structured Active Contours End-To-End}.
\newblock In {\em Conference on Computer Vision and Pattern Recognition}, 2018.

\bibitem{Mattyus17}
G.~{M{\'a}ttyus}, W.~{Luo}, and R.~{Urtasun}.
\newblock {Deeproadmapper: Extracting Road Topology from Aerial Images}.
\newblock In {\em International Conference on Computer Vision}, pages
  3458--3466, 2017.

\bibitem{Min19}
S.~Min, X.~Chen, Z.~Zha, F.~Wu, and Y.~Zhang.
\newblock {A Two-Stream Mutual Attention Network for Semi-Supervised Biomedical
  Segmentation with Noisy Labels}.
\newblock In {\em AAAI Conference on Artificial Intelligence}, pages
  4578--4585, 2019.

\bibitem{Mnih13}
V.~Mnih.
\newblock {\em {Machine Learning for Aerial Image Labeling}}.
\newblock PhD thesis, University of Toronto, 2013.

\bibitem{Mnih10}
V.~Mnih and G.E. Hinton.
\newblock {Learning to Detect Roads in High-Resolution Aerial Images}.
\newblock In {\em European Conference on Computer Vision}, pages 210--223,
  2010.

\bibitem{Mosinska20}
A.~Mosi{\'n}ska, M.~Kozinski, and P.~Fua.
\newblock {Joint Segmentation and Path Classification of Curvilinear
  Structures}.
\newblock {\em IEEE Transactions on Pattern Analysis and Machine Intelligence},
  42(6):1515--1521, 2020.

\bibitem{Mosinska18}
A.~Mosi{\'n}ska, P.~Marquez-Neila, M.~Kozinski, and P.~Fua.
\newblock {Beyond the Pixel-Wise Loss for Topology-Aware Delineation}.
\newblock In {\em Conference on Computer Vision and Pattern Recognition}, pages
  3136--3145, 2018.

\bibitem{Oner21a}
D.~Oner, M.~Kozi{\'n}ski, L.~Citraro, N.~C. Dadap, A.~G. Konings, and P.~Fua.
\newblock {Promoting Connectivity of Network-Like Structures by Enforcing
  Region Separation}.
\newblock {\em IEEE Transactions on Pattern Analysis and Machine Intelligence},
  2021.

\bibitem{Peng14}
H.~Peng, J.~Tang, H.~Xiao, A.~Bria, J.~Zhou, V.~Butler, Z.~Zhou, P.T.
  Gonzalez-Bellido, S.W. Oh, and C.~A. others.
\newblock {Virtual Finger Boosts Three-Dimensional Imaging and Microsurgery as
  Well as Terabyte Volume Image Visualization and Analysis}.
\newblock {\em Nature Communications}, 5:4342--4355, 2014.

\bibitem{Peng17}
H.~Peng, Z.~Zhou, E.Meijering, T.Zhao, G.A. Ascoli, and M.Hawrylycz.
\newblock {Automatic Tracing of Ultra-Volumes of Neuronal Images}.
\newblock {\em Nature Methods}, 14:332--333, 2017.

\bibitem{Ramnath08}
K.~Ramnath, S.~Baker, I.~Matthews, and D.~Ramanan.
\newblock {Increasing the Density of Active Appearance Models}.
\newblock In {\em Conference on Computer Vision and Pattern Recognition}, 2008.

\bibitem{Ronneberger15}
O.~Ronneberger, P.~Fischer, and T.~Brox.
\newblock {{U-Net}: Convolutional Networks for Biomedical Image Segmentation}.
\newblock In {\em Conference on Medical Image Computing and Computer Assisted
  Intervention}, pages 234--241, 2015.

\bibitem{Seyedhosseini13}
M.~Seyedhosseini, M.~Sajjadi, and T.~Tasdizen.
\newblock {Image Segmentation with Cascaded Hierarchical Models and Logistic
  Disjunctive Normal Networks}.
\newblock In {\em International Conference on Computer Vision}, 2013.

\bibitem{Sironi16a}
A.~Sironi, E.~Turetken, V.~Lepetit, and P.~Fua.
\newblock {Multiscale Centerline Detection}.
\newblock {\em IEEE Transactions on Pattern Analysis and Machine Intelligence},
  38(7):1327--1341, 2016.

\bibitem{Terzopoulos88}
D.~Terzopoulos, A.~Witkin, and M.~Kass.
\newblock {Constraints on Deformable Models: Recovering 3D Shape and Nonrigid
  Motion}.
\newblock {\em Artificial Intelligence}, 36(1):91--123, 1988.

\bibitem{Turetken13c}
E.~Turetken, C.~Becker, P.~Glowacki, F.~Benmansour, and P.~Fua.
\newblock {Detecting Irregular Curvilinear Structures in Gray Scale and Color
  Imagery Using Multi-Directional Oriented Flux}.
\newblock In {\em International Conference on Computer Vision}, pages
  1553--1560, December 2013.

\bibitem{APLS}
A.~{Van Etten}.
\newblock {Spacenet Road Detection and Routing Challenge Part II --- APLS
  Implementation}.

\bibitem{Wang20i}
G.~Wang, X.~Liu, C.~Li, Z.~Xu, J.~Ruan, H.~Zhu, T.~Meng, K.~Li, N.~Huang, and
  S.~Zhang.
\newblock {A Noise-Robust Framework for Automatic Segmentation of COVID-19
  Pneumonia Lesions from CT Images}.
\newblock {\em IEEE Transactions on Medical Imaging}, 39(8):2653--2663, 2020.

\bibitem{Wang19c}
W.~Wang, K.~Yu, J.~Hugonot, P.~Fua, and M.~Salzmann.
\newblock {Recurrent U-Net for Resource-Constrained Segmentation}.
\newblock In {\em International Conference on Computer Vision}, 2019.

\bibitem{Wegner13}
J.D. Wegner, J.A. Montoya-Zegarra, and K.~Schindler.
\newblock {A Higher-Order CRF Model for Road Network Extraction}.
\newblock In {\em Conference on Computer Vision and Pattern Recognition}, pages
  1698--1705, 2013.

\bibitem{Wiedemann98}
C.~Wiedemann, C.~Heipke, H.~Mayer, and O.~Jamet.
\newblock {Empirical Evaluation of Automatically Extracted Road Axes}.
\newblock In {\em Empirical Evaluation Techniques in Computer Vision}, pages
  172--187, 1998.

\bibitem{Wu12a}
D.~Wu, D.~Liu, Z.~Puskas, C.~Lu, A.~Wimmer, C.~Tietjen, G.~Soza, and S.~K.
  Zhou.
\newblock {A Learning Based Deformable Template Matching Method for Automatic
  Rib Centerline Extraction and Labeling in CT Images}.
\newblock In {\em Conference on Computer Vision and Pattern Recognition}, 2012.

\bibitem{Yu18d}
Z.~Yu, W.~Liu, Y.~Zou, C.~Feng, S.~Ramalingam, K.~Vijaya, and J.~Kautz.
\newblock {Simultaneous Edge Alignment and Learning}.
\newblock In {\em European Conference on Computer Vision}, 2018.

\bibitem{Zhang19}
Z.~Zhang, D.~Marin, E.~Chesakov, M.~M. Maza, M.~Drangova, and Y.~Boykov.
\newblock {Divergence Prior and Vessel-Tree Reconstruction}.
\newblock In {\em Conference on Computer Vision and Pattern Recognition}, pages
  10216--10224, 2019.

\bibitem{Zhou16}
Z.~Zhou, X.~Liu, B.~Long, and H.~Peng.
\newblock {TReMAP: Automatic 3D Neuron Reconstruction Based on Tracing, Reverse
  Mapping and Assembling of 2D Projections}.
\newblock {\em Neuroinformatics}, 14(1):41--50, January 2016.

\bibitem{Zhu19}
H.~Zhu, J.~Shi, and J.~Wu.
\newblock {Pick-And-Learn: Automatic Quality Evaluation for Noisy-Labeled Image
  Segmentation}.
\newblock In {\em Conference on Medical Image Computing and Computer Assisted
  Intervention}, pages 576--584, 2019.

\end{thebibliography}

\end{document}